\documentclass[gmd, manuscript]{copernicus}

\begin{document}
\nolinenumbers
\title{Using Probabilistic Machine Learning to Better Model Temporal Patterns in Parameterizations: a case study with the Lorenz 96 model.}


\Author[1,2]{Raghul}{Parthipan}
\Author[3]{Hannah M.}{Christensen}
\Author[2,4]{J. Scott}{Hosking}
\Author[1]{Damon J.}{Wischik}

\affil[1]{Department of Computer Science, University of Cambridge, Cambridge, UK}
\affil[2]{British Antarctic Survey, Cambridge, UK}
\affil[3]{Department of Physics, University of Oxford, Oxford, UK}
\affil[4]{The Alan Turing Institute, London, UK}




\correspondence{Raghul Parthipan (rp542@cam.ac.uk)}

\runningtitle{TEXT}

\runningauthor{TEXT}

\received{}
\pubdiscuss{} 
\revised{}
\accepted{}
\published{}


\firstpage{1}

\maketitle

\begin{abstract}
The modelling of small-scale processes is a major source of error in climate models, hindering the accuracy of low-cost models which must approximate such processes through parameterization. Red noise is essential to many operational parameterization schemes, helping model temporal correlations. We show how to build on the successes of red noise by combining the known benefits of stochasticity with machine learning. This is done using a physically-informed recurrent neural network within a probabilistic framework. Our model is competitive and often superior to both a bespoke baseline and an existing probabilistic machine learning approach (GAN) when applied to the Lorenz 96 atmospheric simulation. This is due to its superior ability to model temporal patterns compared to standard first-order autoregressive schemes. It also generalises to unseen scenarios. We evaluate across a number of metrics from the literature, and also discuss the benefits of using the probabilistic metric of hold-out likelihood.

\end{abstract}


\introduction  
\label{chapter_intro}
A major source of inaccuracies in climate models is due to `unresolved' processes. These occur at scales smaller than the resolution of the climate model (`sub-grid' scales) but still have key effects on the overall climate. In fact, most of the inter-model spread in how much global surface temperatures increase after $\textrm{CO}{_2}$ concentrations double is due to the representation of clouds \citep{schneider_clouds,zelinka2020causes}. The typical approach to deal with the problem of unresolved processes has been to model the \textit{effects} of these unresolved processes as a function of the resolved ones. This is known as `parameterization'. 

Red noise is a key feature in many parameterization schemes \citep{christensen_regimes,johnson2019seas5,molteni2011new,palmer2009stochastic,skamarock2019description,walters2019met}. \citet{hasselmann1976stochastic} developed the theory underpinning this, showing that the coupling of processes with different time-scales leads to red noise signals in the longer time-scale process, analogous to what takes place in Brownian motion. In parameterization, the resolved processes typically have far longer time-scales than the unresolved ones, motivating the inclusion of red noise in these schemes. The usefulness of tracking the history of the system for parameterization follows from the importance of red noise.

We use probabilistic machine learning to propose a data-driven successor to red noise in parameterization. The theory explaining the prevalence of red noise in the climate was developed in an idealized model, within the framework of physics-based differential equations. Such approaches may be intractable outside of the idealized case. Recently, there has been much work looking at uncovering relationships from \textit{data} instead \citep{arcomano2022hybrid,beucler2020towards,beucler2021enforcing,bolton2019applications,brenowitz2018prognostic,brenowitz2019spatially,chattopadhyay2020data,gan_hannah,gentine,krasnopolsky,ogorman_2018,rasp2018deep,vlachas2022multiscale,yuval_rf,yuval2021}. We develop a Recurrent Neural Network in a probabilistic framework, combining the benefits of stochasticity with the ability to learn temporal patterns in more flexible ways than permitted by red noise. We use the Lorenz 96 model \citep{lorenz1996predictability}, henceforth the L96, as a proof-of-concept. Our model is competitive with, and often outperforms, a bespoke baseline. It also generalises to unseen scenarios.

\subsection{Numerical Models}

Numerical models represent the state of the Earth system at time $t$ by a state vector $\mathbf{X}_t$. The components of $\mathbf{X}_t$ may include, for example, the mean temperature and humidity at time $t$ in every cell of a grid that covers the Earth. The goal is to parameterize the effects of unresolved processes in such a way that the model can reproduce the evolution of this finite-dimensional representation of the state of the real Earth system. A simple way to model the evolution would be
\begin{linenomath*}
\postdisplaypenalty=0
\begin{align}
    X_{k,t+1} &= X_{k,t} + f_k(\mathbf{X}_t) \label{eq:model0}
\end{align}
\end{linenomath*}
where $X_{k,t}$ is the value that the state variable $X$ takes at the spatial coordinate $k$ and time point $t$ and $X_{k,t} \in \mathbb{R}^d$, $\mathbf{X}_t \in \mathbb{R}^{dK}$, and $f$ is a function for the updating process.

\subsection{Stochastic Schemes}

Stochasticity can be included using the following form
\begin{linenomath*}
\postdisplaypenalty=0
\begin{align}
    X_{k,t+1} &= X_{k,t} + f_k(\mathbf{X}_t,h_{k,t+1}) \label{eq:model1} \\
    h_{k,t+1} &= \beta(h_{k,t},z_{k,t+1}) \label{eq:hidden} 
    \\
    z_{k,t} &\sim \mathcal{N}(0,I)     \label{eq:noise} 
\end{align}
\end{linenomath*}
where invented hidden variables (discussed below) are denoted $h_{k,t} \in \mathbb{R}^H$ and are tracked through time, $\beta$ is a function for updating these, and $z_{k,t} \in \mathbb{R}^Z$ is a source of stochasticity.

Introducing stochasticity through \eqref{eq:model1} improved on models of the form \eqref{eq:model0}. Results included better ensemble forecasts \citep{buizza1999stochastic,leutbecher2017stochastic,palmer2012stochastic}, and improvements to model mean state \citep{berner2012systematic} and climate variability \citep{christensen2017stochastic}. The motivation for using stochasticity comes from the understanding that the effects of the unresolved (sub-grid) processes cannot be effectively predicted as a deterministic function of the resolved ones due to a lack of scale separation between them. Stochasticity allows us to capture our uncertainty about those aspects of the unresolved processes which may affect the resolved outcomes.

\subsection{Hidden Variables and Red Noise}

Hidden variables --- $h_{k,t}$ in~\eqref{eq:model1} and \eqref{eq:hidden} --- are defined here as being variables separate to the observed state, $X_{k,t}$, which if tracked help better model $X_{k,t}$. Using them is key to numerical weather and climate models using stochastic parameterizations, allowing temporal correlations to be better modelled. Red noise results when the hidden variables evolve with a first-order autoregressive (AR1) process such as
\begin{linenomath*}
\begin{equation}
    h_{k,t+1} =\phi h_{k,t} +  \sigma (1-\phi^2)^{1/2}z_{k,t+1} \label{eq:red}
\end{equation}
\end{linenomath*}
One example is the stochastically perturbed parameterization tendencies (SPPT) scheme \citep{buizza1999stochastic,palmer2009stochastic} which is widely used in forecasting models \citep{leutbecher2017stochastic,molteni2011new,palmer2009stochastic,sanchez2016improved,skamarock2019description,stockdale2011ecmwf}. Here, the AR1 process results in far better weather and climate forecasting skill than a simple white noise model, with good modelling of regime behaviour requiring correlated noise \citep{christensen_regimes,dawson2015simulating}. 

There is no intrinsic reason why an AR1 process is the best way to deal with these correlations. It is simply a modelling choice.



\subsection{Machine Learning for Parameterization}

Learning the parameters of a climate model, either the simple form~\eqref{eq:model0} or the general form~\eqref{eq:model1}--\eqref{eq:noise}, requires deciding on a parametric form for the functions $f$ and $\beta$. For full-scale climate models, it is difficult to find appropriate functions. The machine learning (ML) approach is to \emph{learn} these from data. Various researchers have proposed ML methods for learning the deterministic model~\eqref{eq:model0} \citep{beucler2020towards,beucler2021enforcing,bolton2019applications,brenowitz2018prognostic,brenowitz2019spatially,gentine,krasnopolsky,ogorman_2018,rasp2018deep,yuval_rf,yuval2021}.

Amongst ML-trained stochastic models \citep{gan_hannah,guillaumin2021stochastic}, various ones with red noise were proposed by \citet{gan_hannah}, using Generative Adversarial Networks (GANs) \citep{gan_goodfellow}. Full details of the architecture are in \citet{gan_hannah} and we will refer to one of their best-performing models (which they call X-sml-r) as the \emph{GAN}. A wider range of generative models can be trained using such an advesarial approach as opposed to maximum likelihood (the standard way to train models in ML, discussed in Section \ref{chapter_methodology}) but such methods are notoriously unstable due to the nature of the minimax loss in training. Although they used ML to learn $f$ in \eqref{eq:model1}, it was not used to learn $\beta$, which they modelled with an AR1 process.

Recurrent Neural Networks (RNNs) are a popular ML tool for modelling temporally correlated data, eliminating the need for update functions to be manually specified. RNNs have had great success in the ML literature in a variety of sequence modelling tasks, including text generation \citep{graves2013generating,sutskever2011generating}, machine translation \citep{sutskever2014sequence} and music generation \citep{eck2002first,mogren2016c}. The state-of-the-art RNNs are gated ones, principally the long short-term memory (LSTM) networks \citep{lstm} and gated recurrent unit (GRU) networks \citep{gru}, which provide major improvements to the standard issue of unstable gradients.

Current parameterization work using ML to model temporal correlations has used deterministic approaches, including deterministic RNNs and echo state networks \citep{arcomano2022hybrid,chattopadhyay2020data,chattopadhyay2020sp,vlachas2018data,vlachas2020backpropagation,vlachas2022multiscale} and found notable success. This suggests such ML approaches are effective ways to model temporal dynamics in physical systems. However, these studies do not incorporate stochasticity. 

Other off-the-shelf models are not obviously suited for the parameterization task. Transformers and attention-based models \citep{vaswani2017attention} perform well for sequences but require all the previous data to be tracked for each simulation step, providing a computational burden which increases simulation cost. Random Forests (RFs) have been used for parameterization \citep{ogorman_2018,yuval_rf} and shown to be stable at run-time (due to predicting averages form the training set) but it is not obvious how they would learn and track hidden variables. 



\subsection{Overview}

The L96 set-up and baselines are presented in Section \ref{chapter_background}. Our model is detailed in Section \ref{chapter_methodology}, with the experiments and results in Section \ref{chapter_results}. This is followed by a discussion on the use of `likelihood' in evaluating probabilistic climate models (Section \ref{chapter_likelihood}), with our conclusions in Section \ref{chapter_conclusion}.

\section{Parameterization in the Lorenz 96}
\label{chapter_background}

We introduce the L96 model here and then present two L96 parameterization models from the literature which help clarify the above discussion and serve as our baselines.

\subsection{Lorenz 96 Model}

We use the two-tier L96 model, a toy model for atmospheric circulation that is extensively used for stochastic parameterization studies \citep{hannah_bespoke,crommelin2008subgrid,gan_hannah,kwasniok2012data,rasp2020}. We use the configuration described in \citet{gan_hannah}. It consists of two scales of variables: a large, low-frequency variable, $X_k$, coupled to small, high-frequency variables, $Y_{j,k}$. These are dimensionless quantities, evolving as follows:
\begin{linenomath*}
\begin{equation*}
    \frac{dX_k}{dt} =\underbrace{ -X_{k-1}(X_{k-2} - X_{k+1})}_{\textrm{advection}} \underbrace{- X_{k}}_{\textrm{diffusion}} \ \underbrace{+ F}_{\textrm{forcing}} - \underbrace{\frac{hc}{b} \sum_{j=J(k-1)+1}^{kJ} Y_j}_{\textrm{coupling}}  \mkern70mu k = 1, ..., K 
\end{equation*}
\begin{equation*}
    \frac{dY_{j,k}}{dt} = \underbrace{-cbY_{j+1,k}(Y_{j+2,k} - Y_{j-1,k})}_{\textrm{advection}} \: \underbrace{- \, cY_{j,k}}_{\textrm{diffusion}} \: \underbrace{ - \frac{hc}{b} X_k}_{\textrm{coupling}}   \mkern150mu j = 1, ..., J 
\end{equation*}
\end{linenomath*}
where in our experiments, the number of $X_k$ variables is $K=8$, and the number of $Y_{j,k}$ variables per $X_{k}$ is $J=32$. The value of the constants are set to $h=1,\:b=10$ and $\:c=10$. These indicate that the fast variable evolves ten times faster than the slow variable and has one-tenth the amplitude. 

The L96 model is good for parameterization work as we can consider the $X_k$ to be coarse processes resolved in both low-resolution and high-resolution simulators, whilst the $Y_{j,k}$ can be considered as those that can only be resolved in high-resolution, computationally expensive simulators. In this study the coupled set of L96 equations are treated as the `truth', and the aim is to learn a good model for the evolution of $X$ alone using this `truth' data. The effects of $Y_{j,k}$ must therefore be parameterized. Success would be if the modelled evolution of $X$  matched that from the truth.

The L96 is also useful as it contains separate persistent dynamical regimes, which change with different values of $F$ \citep{christensen_regimes,lorenz2006regimes}. The dominant regime exhibits a wave-two pattern around the ring of $X$ variables, whilst the rarer regime exhibits a wave-one type pattern. In the atmosphere, regimes include persistent circulation patterns like the Pacific/North American (PNA) pattern and the North Atlantic Oscillation (NAO). 

It is easy to use models with no memory which do well on standard loss metrics but fail to capture this interesting regime behaviour. This is seen in the L96, where including temporally correlated (red) noise improves the statistics describing regime persistence and frequency of occurrence \citep{christensen_regimes} as well as improving weather forecasting skill \citep{hannah_bespoke,palmer2012stochastic}. Temporally correlated noise also improves regime behaviour (frequency of occurrence and persistence) in operational models \citep{dawson2015simulating}, as well as forecasting skill \citep{palmer2009stochastic}.

\subsection{Parameterization Models in the Literature}

The below models are stochastic. They use AR1 processes to model temporal correlations, but this need not be the case, as we show in Section \ref{chapter_methodology}.

\subsubsection{Stochastic Non-ML Model with Non-ML Hidden Variables}

\citet{hannah_bespoke} propose a model which we refer to as the \emph{polynomial} model (in light of the polynomial in~\eqref{eq:poly}) which is
\begin{linenomath*}
\postdisplaypenalty=0
\begin{align}
   X_{k,t+1} &= X_{k,t} + \omega_k(\mathbf{X}_t) - \Delta t\,\Bigl(a X_{k,t}^3 + b X_{k,t}^2 + c X_{k,t} + d + h_{k,t+1} \Bigr) \label{eq:poly}
\end{align}
\end{linenomath*}
where $h_{k,t}$ evolves as in \eqref{eq:red} with $h_{k,1} = \sigma z_{k,1}$, and where 
\begin{linenomath*}
    \begin{equation}
    \omega_k(\mathbf{X}_t) = \lambda_k\Big(\mathbf{X}_t + \lambda(\mathbf{X}_t/2)\Big)
    \label{eq:rk} \end{equation} 
\end{linenomath*}
and
\begin{linenomath*}
    \begin{equation}
    \lambda_k(\mathbf{X}_t) = \Delta t\,\Big(-X_{k-1,t}(X_{k-2,t} - X_{k+1,t}) - X_{k,t} + F\Big)
    \label{eq:lambda}
    \end{equation}
\end{linenomath*}
where \eqref{eq:rk} is the implementation of a second order Runge-Kutta method and $[\lambda(\mathbf{a})]_k=\lambda_k(\mathbf{a})$. $h_{k,t}$ only depends on $h_{k,t-1}$.

\subsubsection{Stochastic ML Model with Non-ML Hidden Variables}

The GAN from \citet{gan_hannah} replaces \eqref{eq:poly} by
\begin{linenomath*}
\postdisplaypenalty=0
\begin{align}
    X_{k,t+1} &= X_{k,t} + \omega_k(\mathbf{X}_t) - \Delta t\ U(X_{k,t},h_{k,t+1} ) \label{gan1}
\end{align}
\end{linenomath*}
where the function $U$ is implemented by a neural network (NN), $\omega$ is defined in \eqref{eq:rk}--\eqref{eq:lambda}, $h_{k,t}$ evolves as in \eqref{eq:red} with $\sigma = 1$, and $h_{k,1} = z_{k,1}$.

\section{Our Proposed RNN Model}
\label{chapter_methodology}

\subsection{Stochastic ML Model with ML Hidden Variables}

Our model, henceforth denoted the \textit{RNN}, follows the general form in \eqref{eq:model1}--\eqref{eq:noise} but splits the hidden variable into two parts: $h_{k,t} = (r_{k,t},l_{k,t})$. The model is
\begin{linenomath*}
\postdisplaypenalty=0
\begin{align}
    X_{k,t+1} &= X_{k,t} + \omega_k(\mathbf{X}_t) - \Delta t \Big(g_\theta(X_{k,t}) + r_{k,t+1} \Big) \label{eq:rnn1}\\
    r_{k,t+1} &= b_\theta(l_{k,t+1}) + \sigma z_{k,t+1} \label{eq:rnn2} \\
    l_{k,t+1} &= s_\theta(l_{k,t},r_{k,t}) \label{eq:rnn3}
\end{align}
\end{linenomath*}
where $g$, $b$ and $s$ are NNs with weights $\theta$, $z_{k,t}$ is exogenous noise as defined in \eqref{eq:noise}, $\omega$ is specified in equation \eqref{eq:rk} above (implements a second order Runge-Kutta method and expressing physical quantities like advection), and $\theta$ and $\sigma$ are parameters to be learnt. The dimensions of the NNs are: $g_\theta:\mathbb{R}^1 \rightarrow \mathbb{R}^1$, $b_\theta:\mathbb{R}^8 \rightarrow \mathbb{R}^1$ and $s_\theta:\mathbb{R}^9 \rightarrow \mathbb{R}^8$. It is structurally based on a Recurrent Neural Network (RNN) as it has hidden variables, the same update procedure is used each step, and $g$, $b$ and $s$ are NNs.

The key insight is that our model allows more flexibility than the standard AR1 processes for expressing temporal correlations. To clarify this, equation \eqref{eq:rnn1} is identical in form to \eqref{eq:poly}, but the hidden state is evolved in a more flexible manner than the AR1 process in \eqref{eq:red}.

\begin{figure*}[t]
\includegraphics[width=12cm]{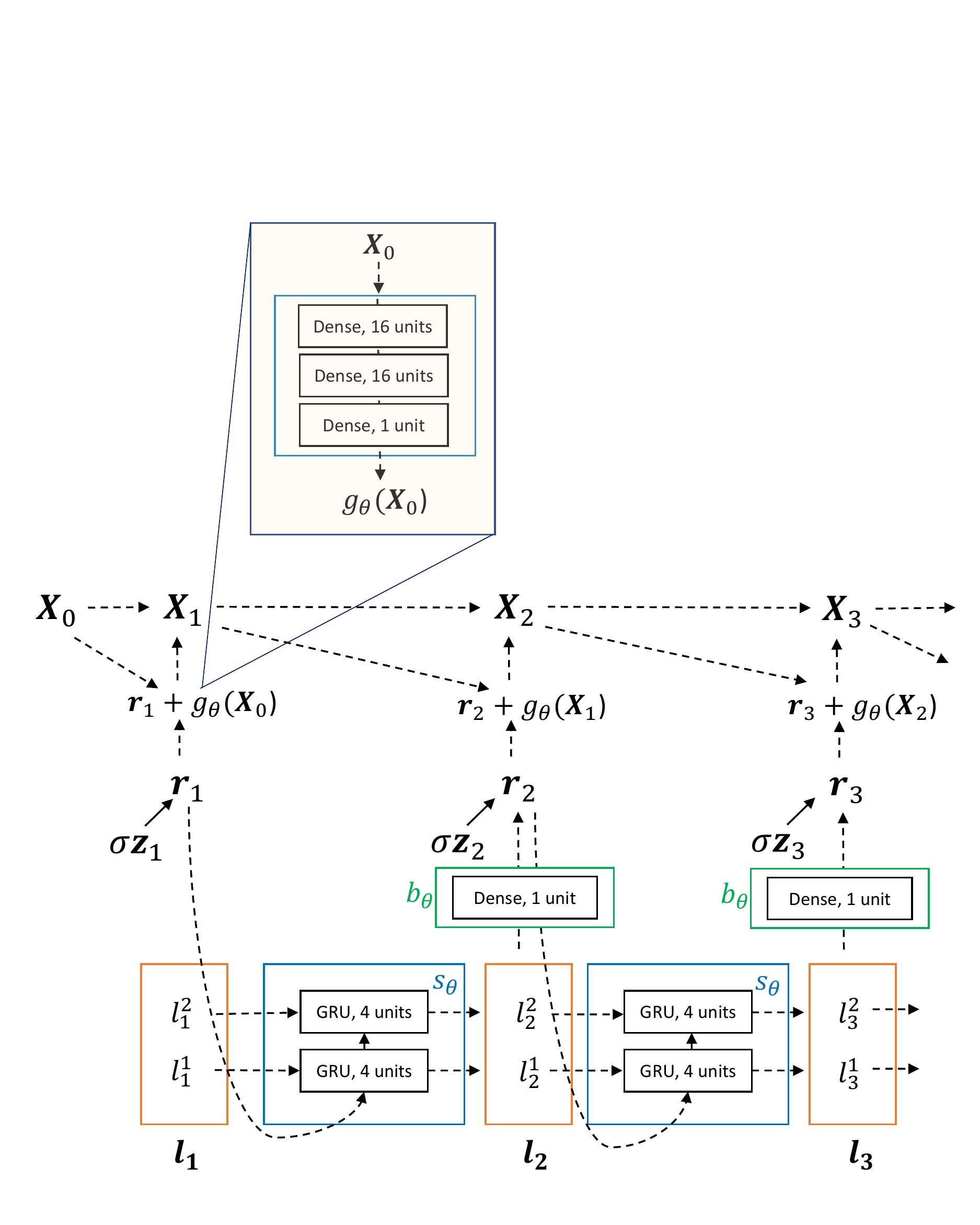}
\caption{RNN graphical model showing how each $\mathbf{X}_{t}$ is generated. $\mathbf{X}_{t+1}$ is a function of $\mathbf{X}_{t}$ and $\mathbf{r}_{t+1}$, and $\mathbf{r}_t$ is a function of $\mathbf{l}_t$ and $\mathbf{z}_t$. $\mathbf{z}_t$ is the source of stochasticity. $b_{\theta}$, $s_{\theta}$ and $g_{\theta}$ are NNs. $\mathbf{l}_t$ consists of a stack of $l_{t}^1$ and $l_{t}^2$, each $\in \mathbb{R}^4$.}
\label{rnn_architecture}
\end{figure*}

Figure \ref{rnn_architecture} shows the mechanism of generation and the NN architectures used to learn the functions. The dotted lines denote the action of deterministic functions, with the solid lines showing the sources of stochasticity. The main architectural details are that $s$ in \eqref{eq:rnn3} is implemented using two GRU layers, each composed of four units, and $b$ in \eqref{eq:rnn2} is represented with a dense layer of size one. $g$ in \eqref{eq:rnn1} is implemented using three fully-connected layers.

Here, as well as in the baselines, the parameterization models are `spatially local' meaning that the full $\mathbf{X}_t$ vector is not taken in as input when modelling $X_{k,t+1}$ anywhere apart from in $\omega_k(\mathbf{X}_t)$. This is done to mimic parameterizations in operational weather and climate models. For all forecast models, $\Delta t = 0.005$ model time units (MTU). One MTU is approximately five atmospheric days when considering the time it takes for errors to double.

\subsection{Training Probability Models Using Likelihood}
\label{chapter_submethod_likelihood}

The RNN is probabilistic and trained using likelihood. The `likelihood of a sequence' is denoted $\mathrm{Pr}(x_1,x_2,...,x_n )$ and can be interpreted as the probability assigned to a given sequence of variables $(x_1,x_2,...,x_n)$ by a given model. 

For our RNN, the log-likelihood (the natural logarithm of the likelihood) of the sequence of $\mathbf{X}_t$ is 
\begin{linenomath*}
\begin{equation*}
    \log\mathrm{Pr}(\mathbf{x}_1,...,\mathbf{x}_n;\mathbf{x}_0) 
 =  \sum_{k=1}^K\Big(\log\mathrm{Pr}(r_{k,1}) + \sum_{t=2}^{n}\log\mathrm{Pr}(r_{k,t}|l_{k,t}) 
  \Big)
 - Kn\log(\Delta t)
\end{equation*}
\end{linenomath*}
where $r$ and $l$ are deterministic functions of $\mathbf{x}_0,...,\mathbf{x}_n$ derived from equations \eqref{eq:rnn1}--\eqref{eq:rnn3}. The full derivation is in \ref{chapter_appendix}.

A common way to train probabilistic models is to maximise the likelihood of the training data. This involves finding a set of parameters ($\theta$ and  $\sigma$ here) which maximise the likelihood. Other enhancements to training involve including regularisation penalties such as dropout. The explicit form of the RNN's likelihood makes training relatively easy.

\subsection{RNN Training}

Truth data for training was created by running the full L96 model five separate times with the following values of $F$: $(19,20,20.5,21,21.5)$. We gave our model data from different forcing scenarios so it could learn how perturbations in the forcing may affect the evolution of $\mathbf{X}_t$. The GAN and polynomial were trained only on $F = 20$ as was done by their authors. The truth data was created by solving the two-level L96 equations using a fourth-order Runge-Kutta timestepping scheme and a fixed time step $\Delta t = 0.001$ MTU, with the output saved at every 0.005 MTU. A training set of length 2,500 MTU was assembled from the truth data, consisting of three equal 500 MTU components with $F = (19,20.5,21)$ and one 1,000 MTU component with $F = 20$, with a $F=21.5$ set of length 500 MTU kept as a validation hold-out set.

The RNN was trained using truncated back propagation through time on sequences of length 700 time steps for 100 epochs with a batch size of 32 using Adam \citep{kingma2014adam}, with $\theta$ and $\sigma$ being the learnable parameters. A variable learning rate was used, starting at 0.0001 for the first 70 epochs and decayed to 0.00003 for the remaining 30. The model parameters which gave the lowest loss on the validation set were saved. 

\section{Results}
\label{chapter_results}

This section analyses performance across a range of time-scales. The first results are for $F = 20$. Assessing the model in the training realm allows us to verify if it can replicate the L96 attractor. Later experiments are for $F \geq 28$. For $F=20$ and $F=28$, 50,000 MTU long simulations ($\approx 685$ `atmospheric years') were created for analysis.

\subsection{Weather Evaluation}

\begin{figure*}[t]
\includegraphics[width=250pt]{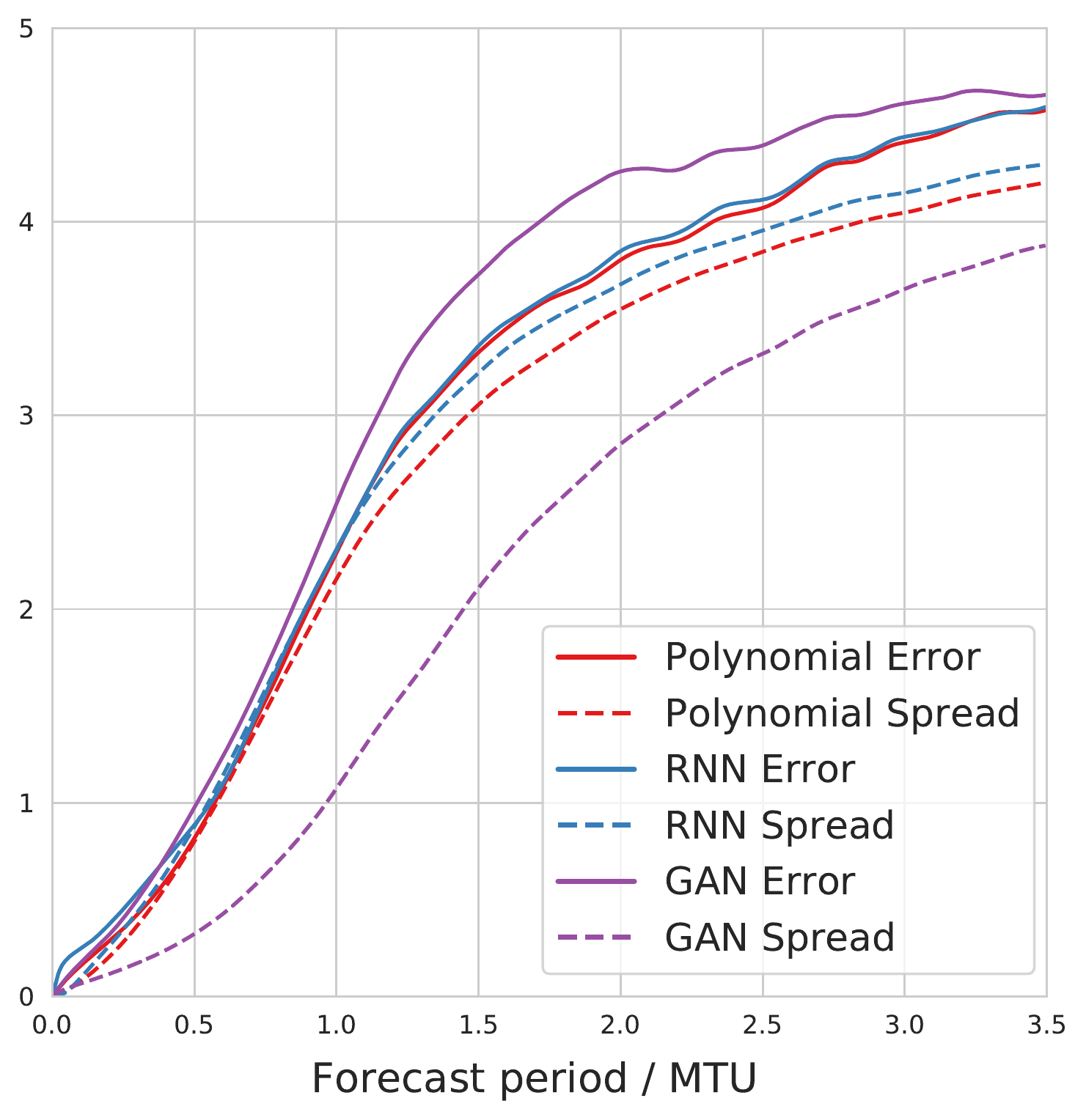}
\caption{Error and spread for weather experiments. We expect a smaller error for a more accurate forecast model. The spread/error ratio would be close to one in a `perfectly reliable' forecast.}
\label{weather_figure}
\end{figure*}

The models were evaluated in a weather forecast framework. 745 initial conditions were randomly selected from the truth attractor and an ensemble of 40 forecasts each lasting 3.5 MTU were produced from each initial condition. Figure \ref{weather_figure} shows the spread and error terms for these experiments over time. The error is defined as
\begin{linenomath*}
\begin{equation*}
    \text{error}(t) = \sqrt{\frac{1}{M}\sum_{m=1}^{M}\big(X_{m}^{O} (t)-\overline{X_{m}^{\text{sample}}(t)}\big)^2}
\end{equation*}
\end{linenomath*}
where there are $M$ different initial conditions, $X_{m}^{O} (t)$ is the observed state at time $t$ for the $m^{\text{th}}$ initial condition, and $\overline{X_{m}^{\text{sample}}(t)}$ is the ensemble mean forecast at time $t$, initialized at $t_{init}$, such that $t = t_{init} + \tau$.

The spread is defined as
\begin{linenomath*}
\begin{equation*}
    \text{spread}(t) = \sqrt{\frac{1}{M}\sum_{m=1}^{M}
    \frac{1}{N}\sum_{n=1}^{N}
    \big(X_{m,n}(t)-
    \overline{X_{m}^{\text{sample}}(t)}\big)^2}
\end{equation*}
\end{linenomath*}
where $N$ is the number of ensemble members and $X_{m,n}(t)$ is the state of the $n^{\text{th}}$ member at time $t$ for the $m^{\text{th}}$ initial condition. 

As noted by \citet{leutbecher2008ensemble}, for a perfectly reliable forecast (defined as one where $X_{m,n}(t)$ and $X_{m}^{O} (t)$ are independent samples from the same distribution), for large $N$ and $M$ the error should be equal to the spread. A reduction in error indicates an ensemble forecast that better tracks observations, and a spread/error ratio close to one indicates a reliable forecast. Figure \ref{weather_figure} shows that the polynomial and RNN have the smallest errors, but the RNN has the better spread/error ratio after 0.75 MTU. The GAN is underdispersive, with the greatest error but smallest spread/error ratio, suggesting it is overconfident in its predictions.

\subsection{Climate Evaluation}

\begin{figure*}[t]
    \includegraphics[width=400pt]{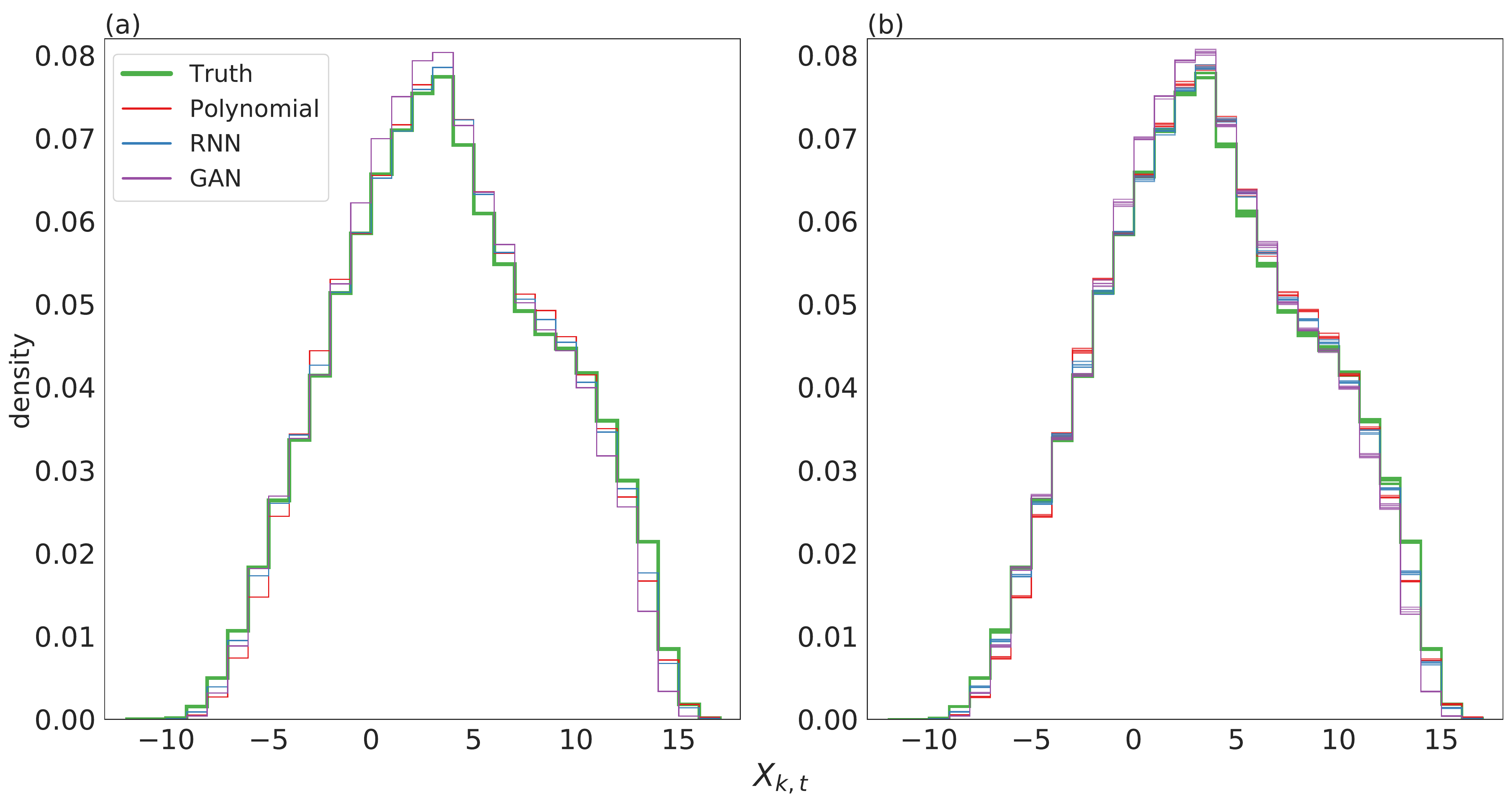}

    \caption{Histograms of $X_{k,t}$. (a) Comparisons for the full simulation data. (b) Depiction which shows sampling error. The truth histogram is shown in green and \textbf{bold}. The closer a model's histogram to the truth, the better it models the probability density of $X_{k,t}$.}
    \label{fig:climate_hist}
\end{figure*}

The ability to simulate the climate of the L96 was evaluated using the 50,000 MTU simulations. The histograms in Figure \ref{fig:climate_hist}a are approximations of each model's marginal distribution of $X_{k,t}$. Successful reproduction of the L96 climate would result in a model's histogram matching the truth model's. Qualitatively, all the models are similar to the truth.

We can quantitatively compare how well histograms match the truth using KL-divergence. We compute this as follows: we discretize our modelled continuous distribution of a variable by binning into a histogram, and do the same for the truth model. We then evaluate the KL-divergence between $q(a)$, the distribution described by the histogram of a variable $A$ in the true L96 model, and $p(a)$, the distribution described by the histogram of that variable in our model
\begin{linenomath*}
 \begin{equation*}
 \text{KL}(q(a)||p(a)) = \sum_{a} q(a) \log \frac{q(a)}{p(a)} 
 \end{equation*}
\end{linenomath*}
where the sum is over all the discrete values which $A$ takes. The smaller this measure, the better the goodness-of-fit.  The results (Table \ref{tab:kl} column one) confirm that the RNN best matches the truth model here.

\begin{table*}
\caption{\textit{KL-divergence (goodness-of-fit) between 1) the truth and 2) the distributions shown in the noted Figures. The smaller the KL-divergence, the better the match between the true and modelled distributions. The best model in each case is shown in \textbf{bold}}.}
\label{tab:kl}
\centering
\begin{tabular}{llllllll}
\hline
Model      & Figure \ref{fig:climate_hist}a        & Figure \ref{fig:regimes}     & Figure \ref{fig:regimes_decomposed}a      & Figure \ref{fig:regimes_decomposed}b               & Figure \ref{fig:regimes_clim_change}a       & Figure \ref{fig:regimes_clim_change}b       \\ \hline
Polynomial & 0.004          & 3.9          & 0.03 & 0.040          & - & -          \\
RNN        & \textbf{0.001} & \textbf{1.8} & \textbf{0.02}          & \textbf{0.004}          & \textbf{0.008}          & 0.006          \\
GAN        & 0.010          & 11.6         & 0.14          & 0.050                & 0.009          & \textbf{0.004}
\end{tabular}
\end{table*}

It it is also important to check whether the histograms are composed of enough data such that they are good representations of each model's actual distribution of $X_{k,t}$. Figure \ref{fig:climate_hist}b seeks to answer this by accounting for sampling error. By splitting a particular model's simulation run into five equally sized sets and plotting these histograms on top of each other, variability can be made apparent. Having done this, there is no noticeable difference between Figure \ref{fig:climate_hist}a and b, suggesting that the error from the sampling process is small relative to the differences between each model.

The L96 model used here displays two distinct regimes with separate dynamics. We use the approach from \citet{christensen_regimes} to examine the regimes. First, the time series are temporally smoothed with a running average over 0.4 MTU to help identify regimes \citep{stephenson2004existence,straus2007circulation}.  The dimensionality of the system is then reduced using Principal Component Analysis, as often done when studying atmospheric data \citep{selten2004preferred,straus2007circulation}. For the truth time series, the components $PC1$ and $PC2$ are degenerate and are in phase quadrature, representing wavenumber two oscillations. $PC3$ and $PC4$ are also degenerate and in phase quadrature, representing wavenumber one oscillations. All model runs are projected onto the truth model's components. Given the degeneracies, the magnitude of the principal component vectors, $||[PC1,PC2]||$ and $||[PC3,PC4]||$ are computed and histograms of the system are plotted in this space. 

\begin{figure*}[t]
    \includegraphics[width=350pt]{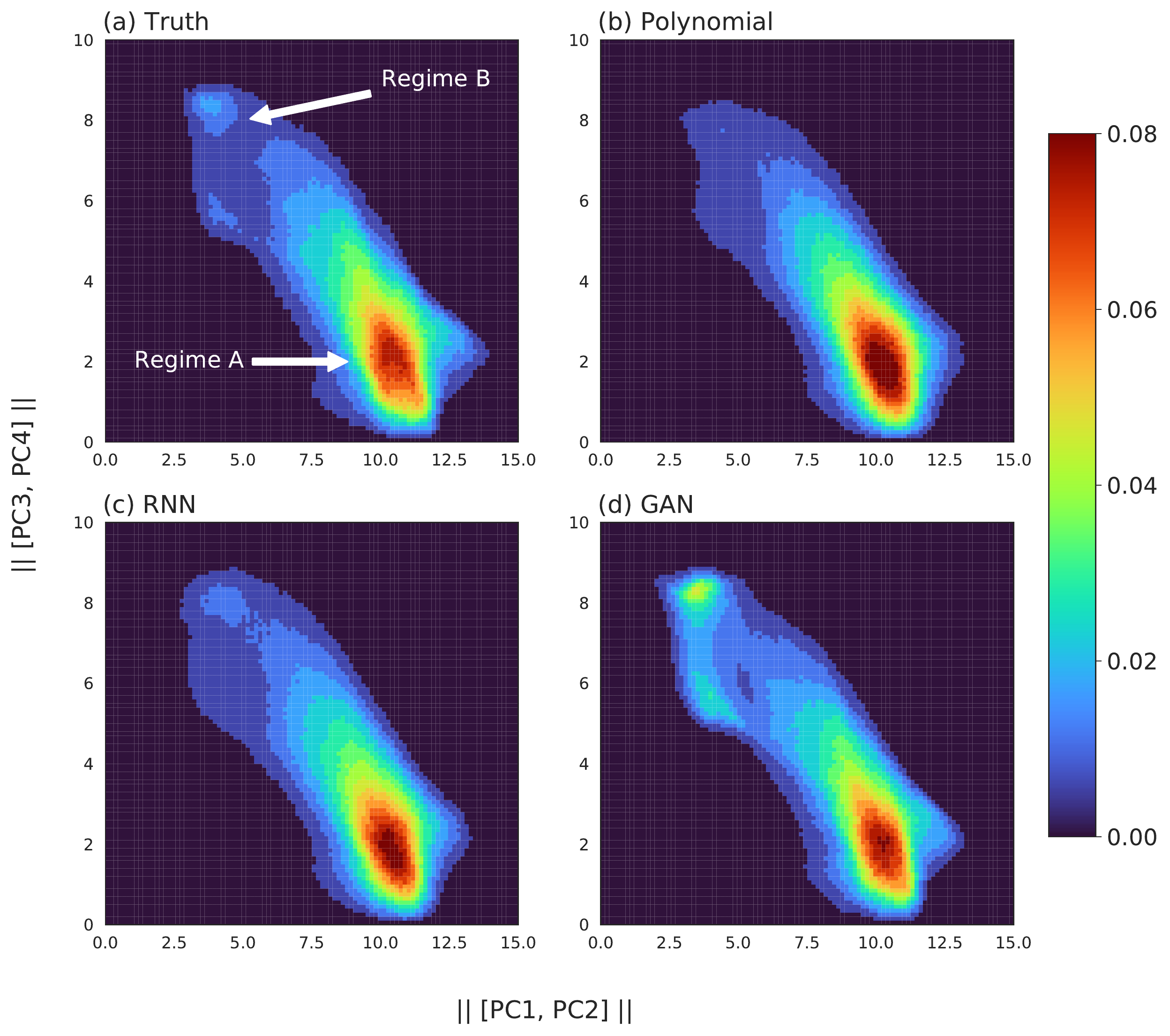}
    \caption{Regime characteristics of the L96. 2D histograms show the magnitudes of the projections of $X_{k,t}$ on to the principle components from the truth series.}
    \label{fig:regimes}
\end{figure*}

\begin{figure*}[t]
    \includegraphics[width=350pt]{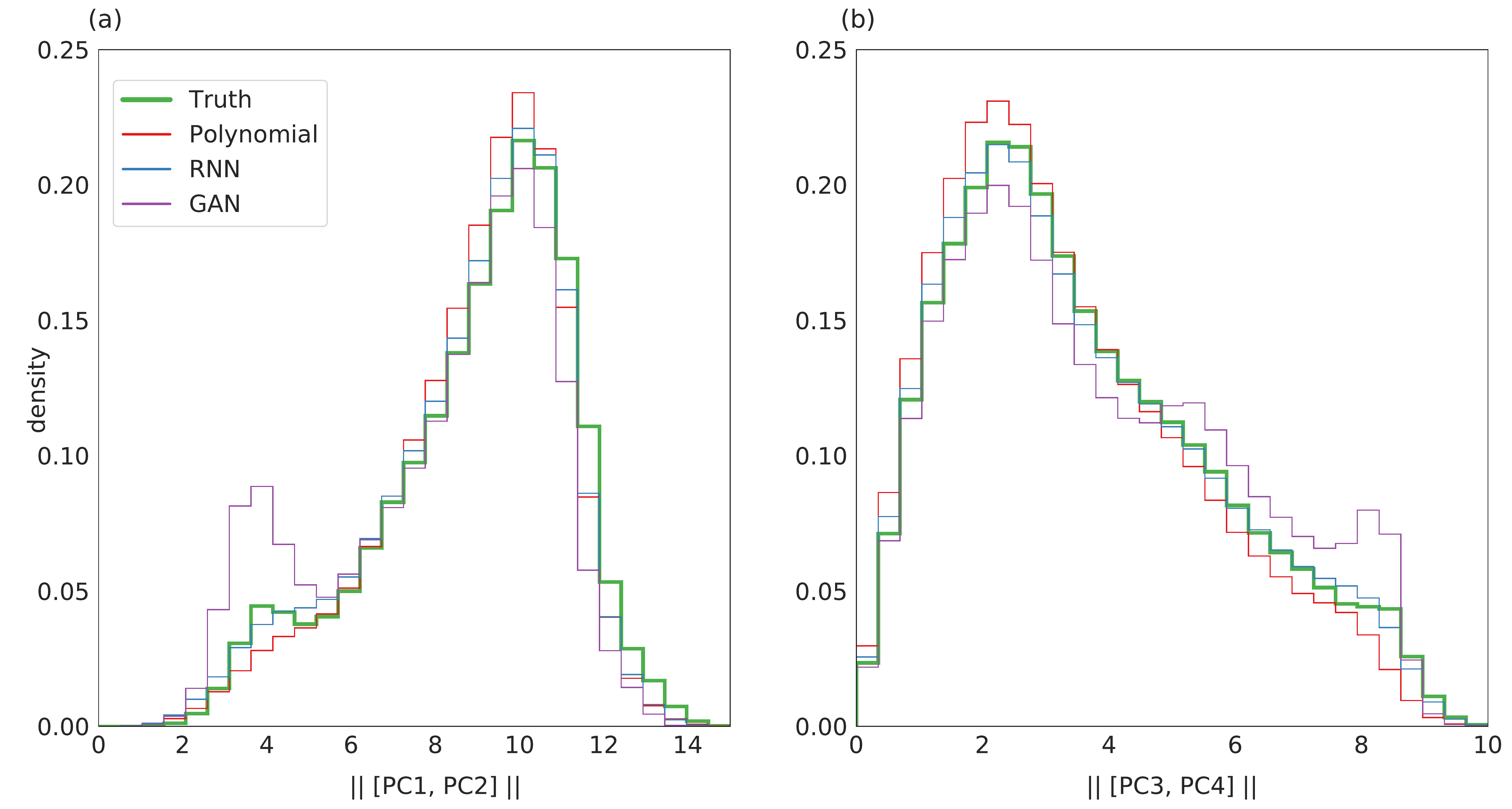}
    \caption{Density plots for each regime for $F=20$. The truth is shown in green and \textbf{bold}. (a) Magnitude of [$PC1$,$PC2$]. (b) Magnitude of [$PC3$,$PC4$]. }
    \label{fig:regimes_decomposed}
\end{figure*}

The presence of two regimes is apparent in Figure \ref{fig:regimes}a where there is a large maximum corresponding to the major regime, A, but also a smaller peak corresponding to the minor regime, B. The polynomial fails to capture the two regimes, whereas the ML models succeed. The RNN puts an appropriate amount of density in Regime B, whilst the GAN puts too much. Comparison is made easier by decomposing the 2D Figure \ref{fig:regimes} into two, 1D density plots (Figure \ref{fig:regimes_decomposed}). We can use the KL-divergence to measure how well our models match the truth across all three histograms. Table \ref{tab:kl} shows the RNN best matches the truth and so captures the regime characteristics best. There are still deficiencies though as it does not put enough density in the right-hand side tail of Figure \ref{fig:regimes_decomposed}a.

\subsection{Generalisation Experiments}

\begin{figure}[t]
    \includegraphics[width=0.5\textwidth]{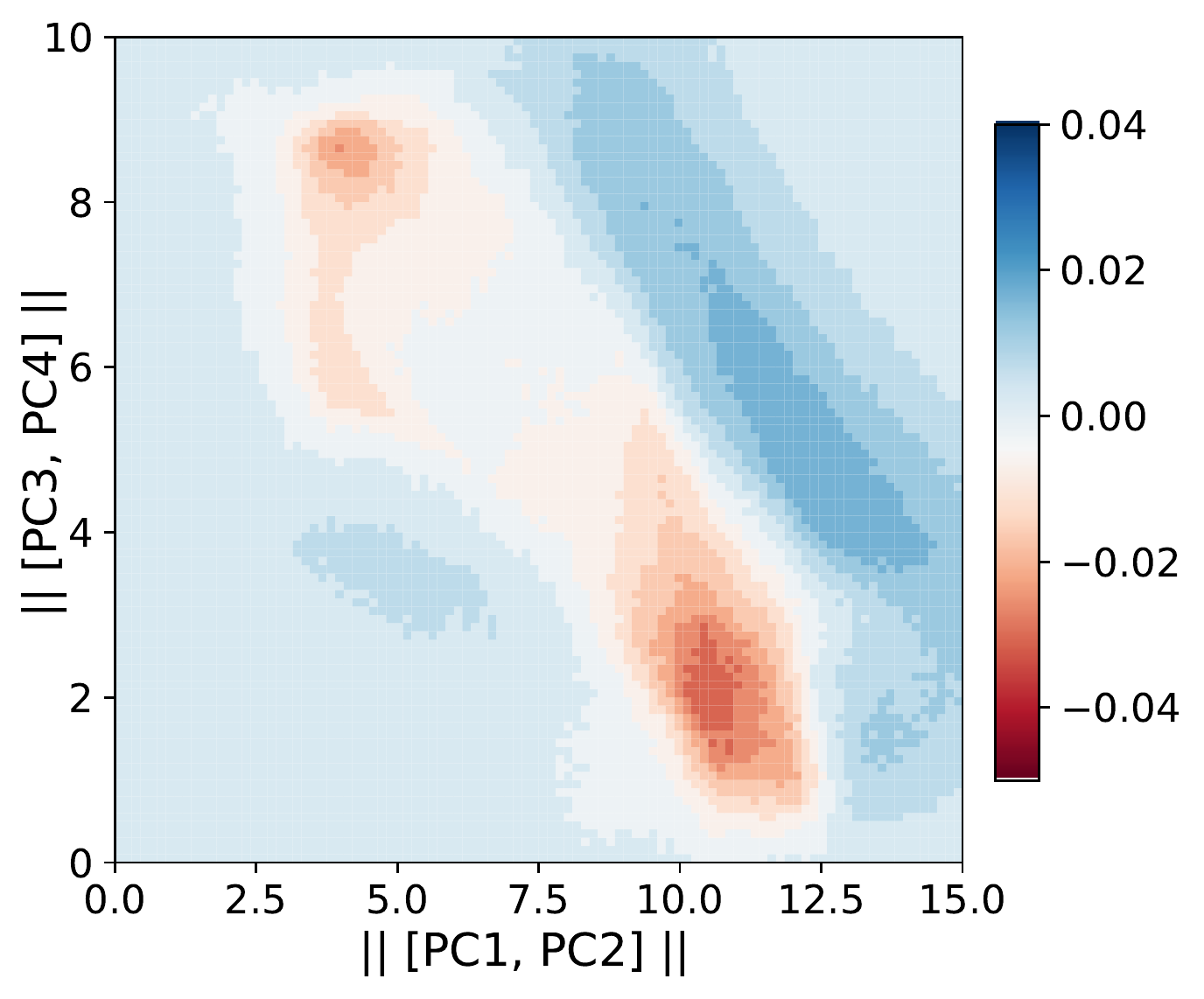}
    \caption{Difference between the $F=21.5$ and $ F = 28$ densities in principal component space. The principal components are those from the $F=20$ truth series.}
    \label{fig:regimes_diff_21_5}
\end{figure}

We set the L96 forcing to $F =28, 32, 35$ and $40$, and examine how the models can capture changes due to varying external forcings. These $F$ values are notably different to those in the training data so allow us to test generalisation. For example, the change from $F = 21.5$ (validation set) to $F=28$ results in the following changes to the regime structure: the centroid location of the rarer regime shifts to higher values of $||[PC1,PC2]||$ and $||[PC3,PC4]||$ (Figure \ref{fig:regimes_diff_21_5}), and the proportion of time spent in the rarer regime increases from $38\%$ to $50\%$. The range of $X_{k,t}$ also increases from $[-12,19]$ to $[-24,29]$. The differences in wave behaviour between regimes means the above changes result in different system dynamics.

In all the below experiments the polynomial model exploded so is omitted. This is merely due to the specification of a third-order polynomial. It significantly deviates from its target values when $X_{k,t}$ values notably different from those in training (such as $X_{k,t} \geq 19$) are taken as inputs. 

\begin{figure*}[t]
    \includegraphics[width=350pt]{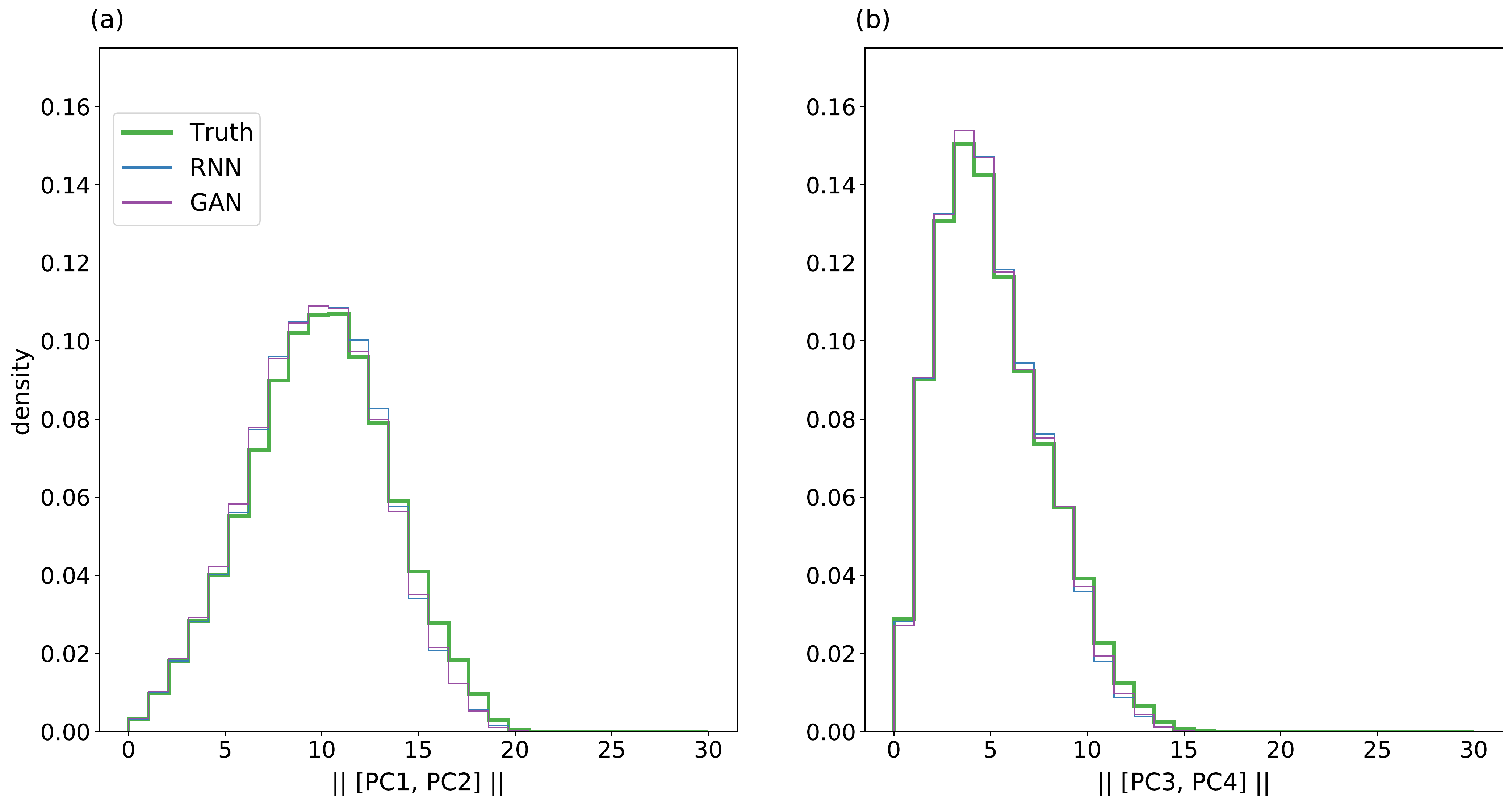}
    \caption{Density plots for each regime for $F=28$. The truth is shown in green and \textbf{bold}. (a) Magnitude of [$PC1$,$PC2$]. (b) Magnitude of [$PC3$,$PC4$]. The right-hand side tails are not giving sufficient density by the RNN nor the GAN.}
    \label{fig:regimes_clim_change}
\end{figure*}

The to simulate the $F = 28$ climate was explored in a similar manner to the $F=20$ one. First, for the histograms of $X_{k,t}$ (analogous to Figure \ref{fig:climate_hist}) the RNN and GAN both had small KL divergences (0.0015 vs 0.0025). Next, the principal component projections were examined (analogous to Figure \ref{fig:regimes}). The same components as determined from the $F=20$ truth data set are used. In this space, the RNN has a smaller KL-divergence (0.6) than the GAN (0.7). Figures for the above two plots are omitted due to it being hard to visually distinguish the models. As before, further comparison can be made by examining the two, 1D density plots in Figure \ref{fig:regimes_clim_change} (analogous to Figure \ref{fig:regimes_decomposed}). Both models perform well (KL divergence in Table \ref{tab:kl}), but neither put enough density in the right-hand side tails.

\begin{figure*}[t]
    \includegraphics[width=350pt]{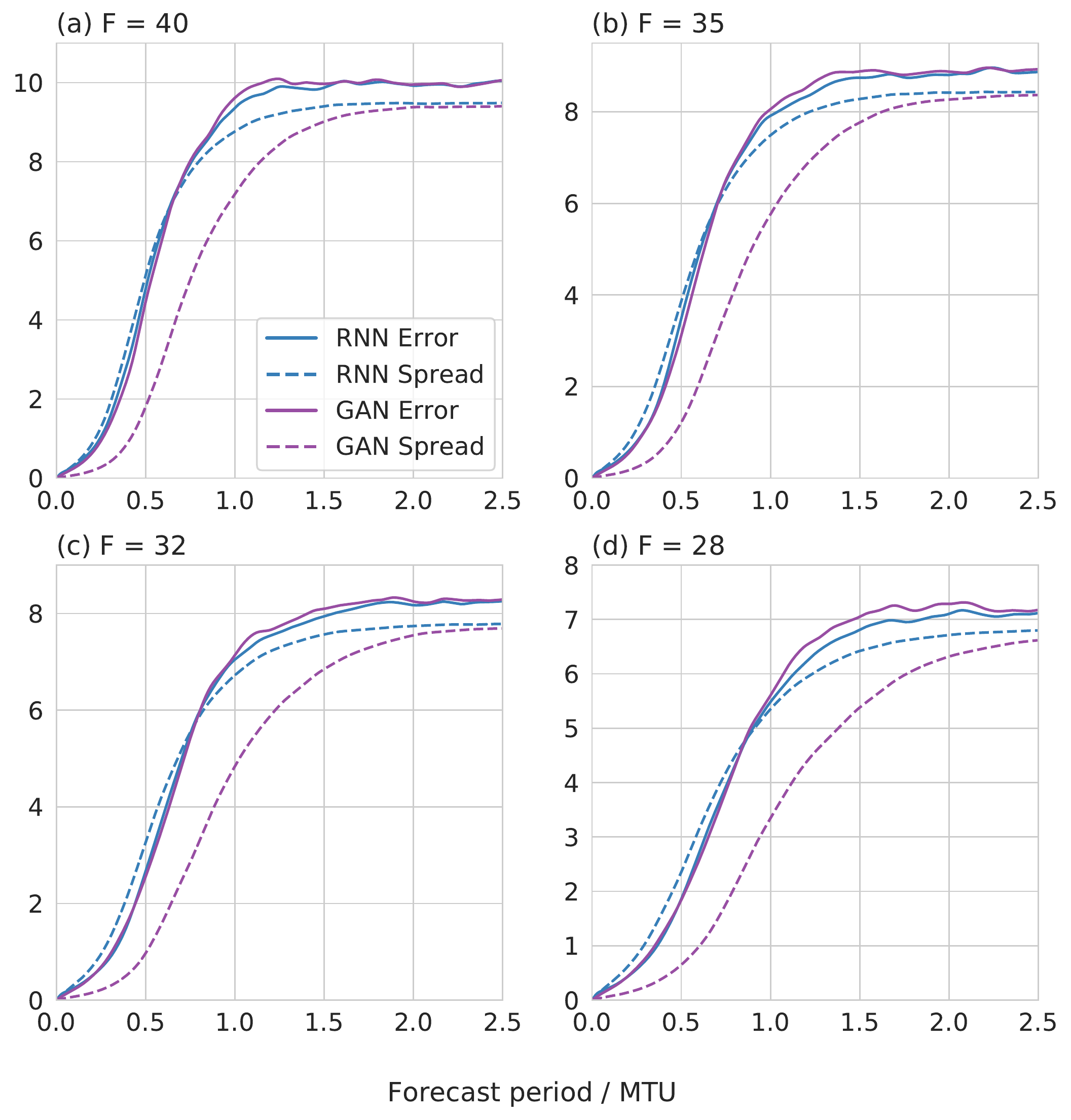}
    \caption{Error and spread for weather experiments for varying forcing values. The RNN's spread is best matched to its error unlike the GAN.}
    \label{fig:weather_generalise}
\end{figure*}

The models were also evaluated in the weather forecasting framework for $F=28, 32, 35$ and $40$. 750 initial conditions were randomly selected from the truth attractors and an ensemble of 40 forecasts each lasting 2.5 MTU were produced. Figure \ref{fig:weather_generalise} shows the RNN and GAN have similar errors, but the RNN's spread is best matched with its error, whereas the GAN continues to be underdispersive.

\subsection{Computational Cost}

We wish our models to have a lower cost than the full L96. The computational cost is calculated by considering the number of floating-point operations per simulated time step ($\Delta t = 0.005$). The RNN  (8,682) is notably cheaper than the truth model (88,000) and the GAN (14,074), so meets the computational cost objective of parameterization work. The polynomial (334) is the cheapest.

\section{Evaluation and Diagnostics Using Likelihood}
\label{chapter_likelihood}

\subsection{Hold-out Likelihood}

We also evaluate using the likelihood (explained in Section \ref{chapter_submethod_likelihood}) of hold-out data. This is a standard approach in the ML literature. Evaluation is done on hold-out sets to ensure that overly complex models which just overfit the training data are not selected. 

Here, hold-out sets for $F=20$ and $F=28$ were created by taking a 10,000 MTU subset from the 50,000 MTU truth sets created in Section \ref{chapter_results}. The hold-out log-likelihoods are shown in Table \ref{tab:loglik_comparisons}. The likelihood for the polynomial model and its derivation is in \ref{chapter_appendix}. We also present an approach to approximate the GAN likelihood. This is despite its full form being intractable (involving integrals which cannot be efficiently approximated using Monte Carlo sampling) and so it not being typically used to evaluate GANs. This is also detailed in \ref{chapter_appendix}. The RNN has a lower likelihood on $F=28$ than the polynomial, despite the polynomial exploding. This is discussed below. 

\subsection{What is the Use of Likelihood}

\subsubsection{For Evaluation}

Other metrics only capture snapshots of model performance. Likelihood is a composite measure which assesses a model's full joint distribution. For example, in the univariate case, the mean-squared-error and variance of a model tell two separate things about its performance, with implicit assumptions being made about variables being normally distributed (whenever mean-squared-error is used). The likelihood captures both of these, and without such assumptions. 

Likelihood also captures information about more complex, joint distributions, saving the need for custom metrics to be invented to assess specific features. To illustrate this, consider we wish to assess a model's temporal correlation. The likelihood already contains this information. Although custom metrics could be invented to assess this, the likelihood is an off-the-shelf metric which is already available. We suggest it is wasteful not to use it.

Being a composite metric brings challenges though. Poor performance in certain aspects may be overshadowed by good performance elsewhere. This can result in cases where increased hold-out likelihoods do not correspond to better sample quality, as noted in the ML literature  \citep{gan_goodfellow,flow_gan,eval_generative_models,zhao2020bridging} and seen in our results: the RNN has a worse $F=28$ hold-out likelihood than the polynomial, yet the polynomial model explodes unlike the RNN. In this case, the phenomenon of `explosion' is not significantly penalising the likelihood. Likelihood is therefore a complement, not replacement, of other metrics which are important to the end-user.

\subsubsection{For Diagnostics}

Likelihood's composite nature makes it a helpful diagnostic tool. Just like KL-divergence, the further away a model is --- in any manner --- from the data in the hold-out set, the worse the likelihood will be. If a model has a poor likelihood despite performing well on a range of standard metrics, this suggests there are still deficiencies in the model which need investigating. For example, with the RNN at $F=28$, the poor average likelihood is caused by a tiny number of segments of the $X_{k,t}$ sequences being extremely poorly modelled (and therefore having extremely poor likelihoods assigned). On inspection, the issue is due to the choice of a fixed $\sigma$ in equation \eqref{eq:rnn2} --- this is appropriate for most of the time series, but for parts which are difficult to model it is too small, preventing the model from expressing sufficient uncertainty. This could be rectified by allowing $\sigma$ to vary, and we suggest future work explore this.

\begin{table}
\caption{\textit{Log-likelihood on hold-out data. The RNN has the best likelihood in both cases.}}
\label{tab:loglik_comparisons}
\centering
\begin{tabular}{lll}
\hline
Model        & $F = 20$       & $F = 28$     \\ \hline
Polynomial   & $4.98$        & $\textbf{4.05}$          \\
RNN & \textbf{6.53} & $2.69$ \\
GAN   & $\approx -37.69 \pm 0.01$            & $\approx -147.02 \pm 0.08$
\end{tabular}
\end{table}

\conclusions[Discussion and Conclusion]  
\label{chapter_conclusion}

We present an approach to replace red noise with a more flexible stochastic machine-learnt model for temporal correlations. Even though we used ML to model $g_\theta(X_{k,t})$ (often called the `sub-grid tendency') in \eqref{eq:rnn1}, the real benefit comes from using ML to model the hidden variables. For example, on setting $g_\theta(X_{k,t})$ in our model equal to $a X_{k,t}^3 + b X_{k,t}^2 + c X_{k,t} + d$ from the polynomial baseline, our model performance hardly suffered (and the cost halved) for $F=20$. This points to a physics-based ML approach where conventional parameterizations are augmented with a stochastic term learnt using ML. 

Using physical knowledge to structure ML models can help with learning. We leveraged many elements from the polynomial baseline, particularly the the physical features, and this gave us better results than when we modelled the system without them. Although in theory a NN can \emph{learn} to create helpful features (such as advection), giving useful features can make learning easier. The NN does not need to learn the known, useful physical relationships and instead can focus on learning other helpful ones. 

There are more sophisticated ways to model the system. We noticed that in some cases the RNN struggled even to overfit the training data, regardless of the RNN complexity. This could be due to difficulties in learning the evolution of the hidden variables. Creating architectures that permit better hidden variables to be learnt (ones which model long-term correlations better) would give better models. Despite our use of GRU cells, which along with the LSTM were great achievements in sequence modelling, we still faced issues with the modelling of long-term trends. Certain models (mainly those which did not leverage physical structure) resulted in unstable simulations. Using hierarchical models which learn to model trends at different temporal resolutions \citep{chung2016hierarchical,liu2020hierarchical} may provide improvement.

\begin{figure*}
    \includegraphics[width=300pt, trim = 1cm 15cm 1.2cm 0.5cm]{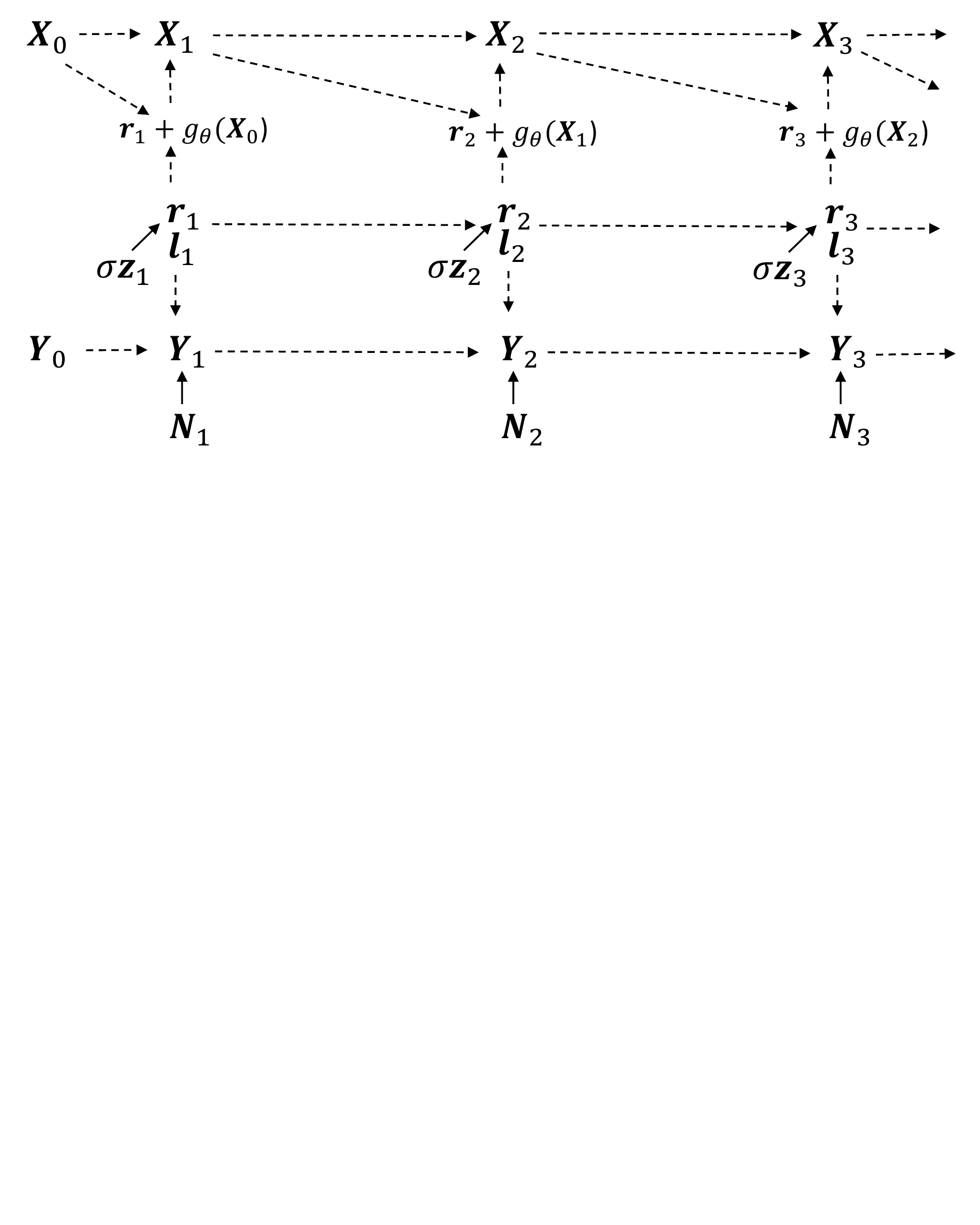}
    \caption{Model which allows learning from high-resolution data without requiring it at simulation time, where $N_t \sim \mathcal{N}(0,I)$.}
    \label{fig:highres}
\end{figure*}

Learning from all the high-resolution data whilst still being computationally efficient at simulation time is an interesting idea. Whilst we cannot keep all the high-resolution dimensions in our schemes (as otherwise we might as well run the high-resolution model outright --- which is not possible as it is too slow for the desired timescales and is a reason why this field of work exists), only keeping the average (as in existing work which coarse-grains data) means much data is thrown away. Our preliminary investigations used the graphical model in Figure \ref{fig:highres}. We designed this so that in training, $\mathbf{l}_t$, is learnt such that it contains useful information about the high-resolution data, $\mathbf{Y}_t$, whereas during generation $\mathbf{Y}_t$ is not required to be simulated. For the L96, this showed no improvement over our RNN though. 

Further work with GANs could result in better models. There may be aspects of realism which we either may not be aware of or may not be able to quantify, so are difficult to include explicitly in our probability models. The GAN discriminator could learn what features constitute `realism' and so the generator may learn to create sequences which contain these. However, our work using GANs and Wasserstein-GANs \citep{arjovsky2017wasserstein,gulrajani2017improved} to train our RNN (instead of by likelihood) did not show benefit. This may be due to the difficulty of training GANs on longer sequences. They have been used to train RNNs on medical timeseries \citep{esteban2017real} and to model music \citep{mogren2016c} so applying this approach to climatic sequences may be something for future work to reconcile. The challenges of using GANs are seen by how we have not been able to reproduce the results shown in \citet{gan_hannah} despite the same set-up. This points to the instability of GAN training as noted by them too. Mode collapse is a common issue encountered when training GANs, where the generator fails to produce samples that explore certain modes of the distribution, and could explain why Regime B in Figure \ref{fig:regimes} has more density than desired. 

\subsection{Conclusion}

We have shown how to build on the benefits of red noise in parameterizations by using a probabilistic ML approach based on an RNN. This now needs testing on more complex systems --- both to specifically improve on AR1 processes, and to learn new, more flexible models of $\mathbf{X}_t$. An interesting area for further work is to examine what information is tracked in the hidden variables of the RNN, and how this relates to the timescales of the required memory in the modelled system. The field is ripe for other probabilistic ML tools to be used and we suspect that further customisation of these will lead to many improved parameterization models.

Likelihood can often be fairly easily calculated, and where this be the case, we propose that the community also evaluate the hold-out likelihood for any devised probabilistic model (ML or otherwise). It is a useful debugging tool, assessing the full joint distribution of a model. It would also provide a consistent evaluation metric across the literature. Given the challenges relating to sample quality, likelihood should complement not replace existing metrics.

Finally, we have used demanding tests to show that ML models can generalise to unseen scenarios. We cannot hope for ML models to generalise to all settings. We do not expect this from our physics-based models either. But ML models \emph{can} generalise outside of their training realm. And it is by using challenging hold-out tests that we can assess their ability to do so, and if they fail, begin the diagnosis of why.




\codedataavailability{All code used in this study (including the code required to create the data) is publicly available at \url{https://github.com/raghul-parthipan/l96_rnn}.} 



\appendix
\section{Likelihood Derivations}   
\label{chapter_appendix}
In all cases, the log-likelihood of the sequence of $\mathbf{X}_t$ is given by
\begin{align}
\begin{split}
    \log\mathrm{Pr}(\mathbf{x}_1,...,\mathbf{x}_n;\mathbf{x}_0) &=  \log\mathrm{Pr}(\mathbf{x}_1;\mathbf{x}_0) + \log\mathrm{Pr}(\mathbf{x}_2|\mathbf{x}_1;\mathbf{x}_0) + \log\mathrm{Pr}(\mathbf{x}_3|\mathbf{x}_2,\mathbf{x}_1;\mathbf{x}_0) \\{}& \qquad \qquad + ... +\log\mathrm{Pr}(\mathbf{x}_n|\mathbf{x}_{n-1},...,\mathbf{x}_1;\mathbf{x}_0) 
\end{split} \label{general_loglik_decomp} 
\end{align}
which follows from the laws of probability.

\subsection{RNN}

Now from \eqref{eq:rnn1}, we can write
\begin{align}
(\mathbf{X}_t|\mathbf{X}_{t-1}{=}\mathbf{x}_{t-1},...,\mathbf{X}_{1}{=}\mathbf{x}_{1};\mathbf{X}_{0}{=}\mathbf{x}_{0}) &= f(\mathbf{x}_{t-1}) - \Delta t  (\mathbf{r}_t|\mathbf{X}_{t-1}{=}\mathbf{x}_{t-1},...,\mathbf{X}_{1}{=}\mathbf{x}_{1};\mathbf{X}_{0}{=}\mathbf{x}_{0})
\end{align}

and given this relationship between $\mathbf{X}_t$ and $\mathbf{r}_t$, a change of variables for the likelihood gives 
\begin{equation}
    \log\mathrm{Pr}(\mathbf{x}_t|\mathbf{x}_{t-1},...,\mathbf{x}_{1};\mathbf{x}_{0}) = \log\mathrm{Pr}(\mathbf{r}_t|\mathbf{r}_{t-1},...,\mathbf{r}_{1}) -K\log(\Delta t) \label{change_of_variables}
\end{equation}
\eqref{general_loglik_decomp} is therefore
\begin{align}
\begin{split} 
     &= \log\mathrm{Pr}(\mathbf{r}_1) +  \log\mathrm{Pr}(\mathbf{r}_2|\mathbf{r}_1)  +  \log\mathrm{Pr}(\mathbf{r}_3|\mathbf{r}_2,\mathbf{r}_1)  \\{}& \qquad \qquad  + ... + \log\mathrm{Pr}(\mathbf{r}_{n}|\mathbf{r}_{n-1:1}) - Kn\log(\Delta t)
\end{split} \label{eq:rnn_deriv_0} \\
&=\log\mathrm{Pr}(\mathbf{r}_1) + \sum_{t=2}^{n}\log\mathrm{Pr}(\mathbf{r}_{t}|\mathbf{l}_{t})  - Kn\log(\Delta t) \label{eq:rnn_deriv_1}\\
&=  \sum_{k=1}^K\Big(\log\mathrm{Pr}(r_{k,1}) + \sum_{t=2}^{n}\log\mathrm{Pr}(r_{k,t}|l_{k,t}) 
  \Big)
 - Kn\log(\Delta t) \label{eq:rnn_deriv_2}
\end{align}
where \eqref{eq:rnn_deriv_1}--\eqref{eq:rnn_deriv_2} follow from the independencies in the graphical model (Figure \ref{rnn_architecture}).

\subsection{Polynomial Model}

For the polynomial model, the log-likelihood of the $\mathbf{X}_t$ sequence is 
\begin{linenomath*}
\begin{align}
    \log\mathrm{Pr}(\mathbf{x}_1,...,\mathbf{x}_n;\mathbf{x}_0) 
 &= \sum_{k=1}^{K}\Big(\log\mathrm{Pr}(h_{k,1})+  \sum_{t=1}^{n-1}\log\mathrm{Pr}(h_{k,t+1}|h_{k,t})\Big) - Kn\log( \Delta t)
\end{align}
\end{linenomath*}
and the derivation follows below.

\begin{figure*}
    \includegraphics[width=300pt, trim = 1cm 7cm 1.2cm 1cm]{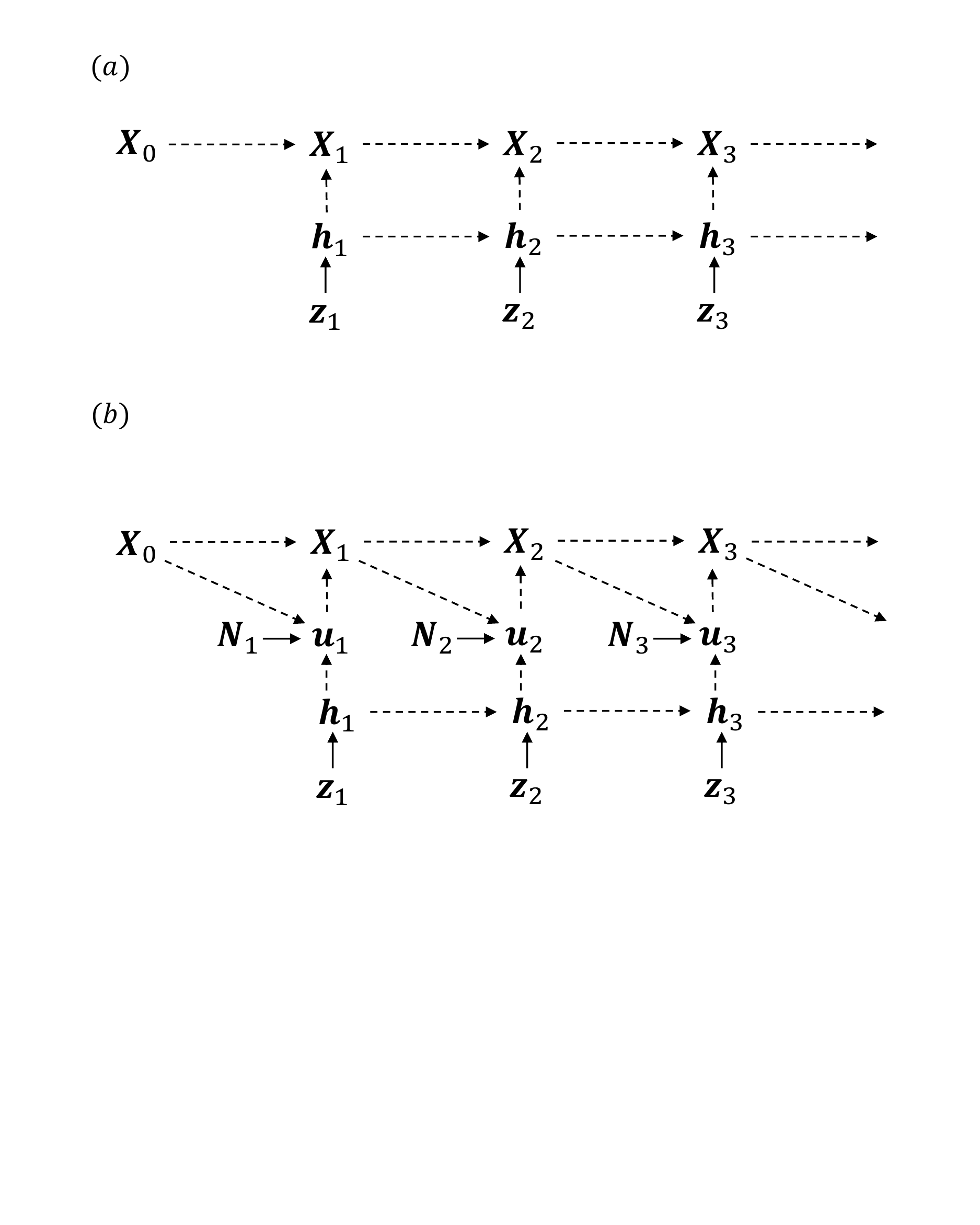}
    \caption{Graphical model for (a) Polynomial and (b) GAN with added white noise.}
    \label{fig:other_graphical_models}
\end{figure*}

The graphical model is shown in Figure \ref{fig:other_graphical_models}a. From \eqref{eq:poly}, we can write
\begin{linenomath*}
\begin{align}
(\mathbf{X}_t|\mathbf{X}_{t-1}{=}\mathbf{x}_{t-1},...,\mathbf{X}_{1}{=}\mathbf{x}_{1};\mathbf{X}_{0}{=}\mathbf{x}_{0}) &= f(\mathbf{x}_{t-1}) - \Delta t  (\mathbf{h}_t|\mathbf{X}_{t-1}{=}\mathbf{x}_{t-1},...,\mathbf{X}_{1}{=}\mathbf{x}_{1};\mathbf{X}_{0}{=}\mathbf{x}_{0})
\end{align}
\end{linenomath*}
and given this relationship between $\mathbf{X}_t$ and $\mathbf{h}_t$, a change of variables for the likelihood gives 
\begin{linenomath*}
\begin{equation}
    \log\mathrm{Pr}(\mathbf{x}_t|\mathbf{x}_{t-1},...,\mathbf{x}_{1};\mathbf{x}_{0}) = \log\mathrm{Pr}(\mathbf{h}_t|\mathbf{h}_{t-1},...,\mathbf{h}_{1}) -K\log(\Delta t) \label{change_of_variables_poly}
\end{equation}
\end{linenomath*}
\eqref{general_loglik_decomp} is therefore \begin{linenomath*}
\begin{align}
\begin{split} 
     &= \log\mathrm{Pr}(\mathbf{h}_1) +  \log\mathrm{Pr}(\mathbf{h}_2|\mathbf{h}_1)  +  \log\mathrm{Pr}(\mathbf{h}_3|\mathbf{h}_2,\mathbf{h}_1)  \\{}& \qquad \qquad  + ... + \log\mathrm{Pr}(\mathbf{h}_{n}|\mathbf{h}_{n-1:1}) - Kn\log(\Delta t)
\end{split} \\
    &= \log\mathrm{Pr}(\mathbf{h}_1) + \sum_{t=1}^{n-1}\log\mathrm{Pr}(\mathbf{h}_{t+1}|\mathbf{h}_{t}) - Kn\log(\Delta t) \label{poly_temp} \\
 &= \sum_{k=1}^{K}\Big(\log\mathrm{Pr}(h_{k,1})+  \sum_{t=1}^{n-1}\log\mathrm{Pr}(h_{k,t+1}|h_{k,t})\Big) - Kn\log(\Delta t) \label{poly_spatial_depend}
\end{align}
\end{linenomath*}
where \eqref{poly_temp}--\eqref{poly_spatial_depend} follow from the independencies described in \eqref{eq:red}.

\subsection{GAN}

We approximate the GAN's likelihood by calculating the likelihood of a model which functions and gives results almost identical to the GAN (one with a small amount of  white noise added) using importance sampling and the reparameterization method as in the Variational Autoencoder \cite{vae}. Here, \eqref{gan1} is altered to
\begin{linenomath*}
\begin{align}
X_{k,t+1} &= X_{k,t} + \omega_k(\mathbf{X}_t) - \Delta t\ u(X_{k,t},h_{k,t+1}) \label{eq:gan3}
\end{align}
\end{linenomath*}
where $u(X_{k,t},h_{k,t+1}) = U(X_{k,t},h_{k,t+1}) + N_{k,t+1}$ and $N_{k,t} \sim \mathcal{N}(0,0.001^2)$. The log-likelihood is 
\begin{linenomath*}
\begin{align}
      \log\mathrm{Pr}(\mathbf{x}_1,...,\mathbf{x}_n;\mathbf{x}_0)   &= \log\mathbb{E}_{\mathbf{h} \sim \mathrm{Pr}_\mathbf{\tilde{H}}.} \frac{\mathrm{Pr}(\mathbf{u}.|\mathbf{h}.,\mathbf{x}_0) \mathrm{Pr}_\mathbf{H.}(\mathbf{h}.)}
    {\mathrm{Pr}_\mathbf{\tilde{H}.}(\mathbf{h}.)}  
    -Kn\log(\Delta t)
    \notag
\end{align}
\end{linenomath*}
where the expectation can now be approximated by a large sum if there is a suitable importance sampler $\mathrm{Pr}_\mathbf{\tilde{H}.}(\mathbf{h}.)$. 50 samples were used for the importance sampling, and this was repeated 50 times to give 95\% confidence intervals. We note that given the finite number of samples taken it is of course possible that the GAN likelihood is larger, but a well-trained importance sampler should minimise the chance of this. The derivation and form of the importance sampler is detailed below.

The graphical model of the GAN with a small amount of white noise added is shown in Figure \ref{fig:other_graphical_models}b. $\mathbf{X}_t$ is related to $\mathbf{u}_{t}$ as in \eqref{eq:gan3} so
\begin{linenomath*}
\begin{align}
(\mathbf{X}_t|\mathbf{X}_{t-1}{=}\mathbf{x}_{t-1},...,\mathbf{X}_{1}{=}\mathbf{x}_{1};\mathbf{X}_{0}{=}\mathbf{x}_{0}) &= f(\mathbf{x}_{t-1}) - \Delta t  (\mathbf{u}_t|\mathbf{X}_{t-1}{=}\mathbf{x}_{t-1},...,\mathbf{X}_{1}{=}\mathbf{x}_{1};\mathbf{X}_{0}{=}\mathbf{x}_{0})
\end{align}
\end{linenomath*}
and a change of variables for likelihood gives
\begin{linenomath*}
\begin{equation}
    \log\mathrm{Pr}(\mathbf{x}_t|\mathbf{x}_{t-1},...,\mathbf{x}_{1};\mathbf{x}_{0}) = \log\mathrm{Pr}(\mathbf{u}_t|\mathbf{u}_{t-1},...,\mathbf{u}_{1};\mathbf{x}_0) -K\log(\Delta t)
\end{equation}
\end{linenomath*}
\eqref{general_loglik_decomp} is therefore
\begin{linenomath*}
\begin{align}
\begin{split} 
     &=  \log\mathrm{Pr}(\mathbf{u}_1|\mathbf{x}_0) +   \log\mathrm{Pr}(\mathbf{u}_2|\mathbf{u}_1,\mathbf{x}_0)  +  \log\mathrm{Pr}(\mathbf{u}_3|\mathbf{u}_2,\mathbf{u}_1,\mathbf{x}_0)  \\{}& \qquad \qquad  + ... + \log\mathrm{Pr}(\mathbf{u}_{n}|\mathbf{u}_{n-1:0},\mathbf{x}_0) -Kn\log(\Delta t)\notag
\end{split} \\ 
    &= \log\mathrm{Pr}(\mathbf{u}.|\mathbf{x}_0) -Kn\log(\Delta t) \notag
\end{align}
\end{linenomath*}
and just decomposing the first term below:
\begin{linenomath*}
\begin{align}
    \log\mathrm{Pr}(\mathbf{u}.|\mathbf{x}_0) &= \log\mathbb{E}_{\mathbf{h} \sim \mathrm{Pr}_\mathbf{H}.}\mathrm{Pr}(\mathbf{u}.|\mathbf{h}.,\mathbf{x}_0) \notag\\
    &= \log\mathbb{E}_{\mathbf{h} \sim \mathrm{Pr}_\mathbf{\tilde{H}}.} \frac{\mathrm{Pr}(\mathbf{u}.|\mathbf{h}.,\mathbf{x}_0) \mathrm{Pr}_\mathbf{H.}(\mathbf{h}.)}
    {\mathrm{Pr}_\mathbf{\tilde{H}.}(\mathbf{h}.)} \label{approx_lik_gan}
\end{align}
\end{linenomath*}
where \eqref{approx_lik_gan} can be approximated with a sufficiently large sum given a good enough encoder $\mathrm{Pr}_\mathbf{\tilde{H}.}(\mathbf{h}.)$. 

For training purposes, the lower-bound is used:
\begin{linenomath*}
\begin{align}
    &\geq \mathbb{E}_{\mathbf{h} \sim \mathrm{Pr}_\mathbf{\tilde{H}}.} \log\frac{\mathrm{Pr}(\mathbf{u}.|\mathbf{h}.,\mathbf{x}_0) \mathrm{Pr}_\mathbf{H.}(\mathbf{h}.)}
    {\mathrm{Pr}_\mathbf{\tilde{H}.}(\mathbf{h}.)} \label{elbo_gan}
\end{align}
\end{linenomath*}

The terms in the numerator decompose as follows, using the independencies from the graphical model and associated equations:
\begin{linenomath*}
\begin{align*}
   \log\mathrm{Pr}(\mathbf{u}.|\mathbf{h}.,\mathbf{x}_0) &=  \sum_{t=1}^{n}\log\mathrm{Pr}(\mathbf{u}_t|\mathbf{h}_t,\mathbf{x}_{t-1}) \\
   &=  \sum_{t=1}^{n}\sum_{k=1}^{K}\log\mathrm{Pr}(u_{k,t}|h_{k,t},x_{k,t-1})
\end{align*}
\end{linenomath*}
and 
\begin{linenomath*}
\begin{align*}
   \log\mathrm{Pr}_\mathbf{H.}(\mathbf{h}.)&= \log \mathrm{Pr}_\mathbf{H}(\mathbf{h}_1) +  \sum_{t=2}^{n}\log\mathrm{Pr}_\mathbf{H}(\mathbf{h}_t|\mathbf{h}_{t-1})  \\
   &= \sum_{k=1}^{K} \Big(\log \mathrm{Pr}_{H}(h_{k,1}) +  \sum_{t=2}^{n}\log\mathrm{Pr}_H(h_{k,t}|h_{k,t-1}) \Big)
\end{align*}
\end{linenomath*}

The perfect importance sampling distribution would be proportional to $\mathrm{Pr}(\mathbf{h}.|\mathbf{u}.,\mathbf{x}_0)$. Therefore we want $\mathrm{Pr}_\mathbf{\tilde{H}.}(\mathbf{h}.)$ to allow for the same dependencies. We choose it by carrying out the following steps:
\begin{linenomath*}
\begin{align}
   \log\mathrm{Pr}_\mathbf{\tilde{H}.}(\mathbf{h}.)&\propto \log\mathrm{Pr}(\mathbf{h}.|\mathbf{u}.,\mathbf{x}_0) \notag \\
   &= \log\mathrm{Pr}_{\mathbf{H}_1}(\mathbf{h}_1|\mathbf{u}.,\mathbf{x}_0) + \sum_{t=2}^{n}\log\mathrm{Pr}_{\mathbf{H}_t}(\mathbf{h}_t|\mathbf{h}_{t-1},\mathbf{u}.,\mathbf{x}_0) \label{is_1}
\end{align}
\end{linenomath*}
where \eqref{is_1} follows from using the laws of probability to decompose the term above, followed by applications of the independencies from the graphical model. Note that although the process is Markov it is not necessarily time-homogeneous. To deal with this, an RNN is used to model $\mathrm{Pr}_{\mathbf{H}_t}(\mathbf{h}_t|\mathbf{h}_{t-1},\mathbf{u}.,\mathbf{x}_0)$ as $\mathrm{Pr}_\mathbf{H}(\mathbf{h}_t|\mathbf{h}_{t-1},\mathbf{u}.,\mathbf{x}_0,\mathbf{s}_t)$ where the state $\mathbf{s}_t$ can in principle account for the non-homogeneous updating procedure of $\mathbf{h}_t$. The log-likelihood is therefore
\begin{linenomath*}
\begin{align}
  &= \sum_{k=1}^K \Big( \log\mathrm{Pr}_{H_1}(h_{k,1}|u_{k.},x_{k,0}) + \sum_{t=2}^{n}\log\mathrm{Pr}_{H_t}(h_{k,t}|h_{k,t-1},u_{k.},x_{k.},s_{k,t}) \Big) \label{is_2}
\end{align}
\end{linenomath*}
where the spatial independencies introduced in \eqref{is_2} follow from the graphical model and $u_{k,.} = u_{k,1:n}$ and $x_{k,.} = x_{k,0:n-1}$. The only simplifying assumption used here is that the same update model for $h_{k,t}$ is used for each component $k$. Finally, the sequence of $(u_{k.},x_{k.})$ is summarised using a bidirectional RNN to give $w_{k,t}$. Therefore, the final form of the log-likelihood of our importance sampler is 
\begin{linenomath*}
\begin{align}
  &= \sum_{k=1}^K \Big( \log\mathrm{Pr}_{H_1}(h_{k,1}|w_{k,1},x_{k,0}) + \sum_{t=2}^{n}\log\mathrm{Pr}_{H_t}(h_{k,t}|h_{k,t-1},w_{k,t}, x_{k,t-1},s_{k,t}) \Big) 
\end{align}
\end{linenomath*}

The training of the importance sampler is done by maximising \eqref{elbo_gan}, with the learnable parameters being those of the model used to learn $h_{k,1}$, the RNN used to model the update of $h_{k,t}$, and the bidirectional RNN.

\noappendix       







\authorcontribution{All authors supported the design of the study. RP created the models and conducted research. RP prepared the manuscript with contributions from all co-authors.} 

\competinginterests{The authors declare that they have no conflict of interest.} 


\begin{acknowledgements}
R.P. was funded by the Engineering and Physical Sciences Research Council [grant number  EP/S022961/1]. H.M.C. was funded by Natural Environment Research Council [grant number NE/P018238/1]. J.S.H. was supported by the British Antarctic Survey, Natural Environment Research Council (NERC) National Capability funding. D.J.W. was funded by the University of Cambridge.
\end{acknowledgements}




\bibliographystyle{copernicus}
\bibliography{bibliography.bib}

\begin{thebibliography}{65}
\providecommand{\natexlab}[1]{#1}
\providecommand{\url}[1]{{\tt #1}}
\providecommand{\urlprefix}{URL }
\expandafter\ifx\csname urlstyle\endcsname\relax
  \providecommand{\doi}[1]{https://doi.org/\discretionary{}{}{}#1}\else
  \providecommand{\doi}{https://doi.org/\discretionary{}{}{}\begingroup
  \urlstyle{rm}\Url}\fi

\bibitem[{Arcomano et~al.(2022)Arcomano, Szunyogh, Wikner, Pathak, Hunt, and
  Ott}]{arcomano2022hybrid}
Arcomano, T., Szunyogh, I., Wikner, A., Pathak, J., Hunt, B.~R., and Ott, E.: A
  Hybrid Approach to Atmospheric Modeling That Combines Machine Learning With a
  Physics-Based Numerical Model, Journal of Advances in Modeling Earth Systems,
  14, e2021MS002\,712, 2022.

\bibitem[{Arjovsky et~al.(2017)Arjovsky, Chintala, and
  Bottou}]{arjovsky2017wasserstein}
Arjovsky, M., Chintala, S., and Bottou, L.: Wasserstein generative adversarial
  networks, in: International conference on machine learning, pp. 214--223,
  PMLR, 2017.

\bibitem[{Arnold et~al.(2013)Arnold, Moroz, and Palmer}]{hannah_bespoke}
Arnold, H.~M., Moroz, I.~M., and Palmer, T.~N.: Stochastic parametrizations and
  model uncertainty in the Lorenz’96 system, Philosophical Transactions of
  the Royal Society A: Mathematical, Physical and Engineering Sciences, 371,
  20110\,479, 2013.

\bibitem[{Berner et~al.(2012)Berner, Jung, and Palmer}]{berner2012systematic}
Berner, J., Jung, T., and Palmer, T.: Systematic model error: The impact of
  increased horizontal resolution versus improved stochastic and deterministic
  parameterizations, Journal of Climate, 25, 4946--4962, 2012.

\bibitem[{Beucler et~al.(2020)Beucler, Pritchard, Gentine, and
  Rasp}]{beucler2020towards}
Beucler, T., Pritchard, M., Gentine, P., and Rasp, S.: Towards
  physically-consistent, data-driven models of convection, in: IGARSS 2020-2020
  IEEE International Geoscience and Remote Sensing Symposium, pp. 3987--3990,
  IEEE, 2020.

\bibitem[{Beucler et~al.(2021)Beucler, Pritchard, Rasp, Ott, Baldi, and
  Gentine}]{beucler2021enforcing}
Beucler, T., Pritchard, M., Rasp, S., Ott, J., Baldi, P., and Gentine, P.:
  Enforcing analytic constraints in neural networks emulating physical systems,
  Physical Review Letters, 126, 098\,302, 2021.

\bibitem[{Bolton and Zanna(2019)}]{bolton2019applications}
Bolton, T. and Zanna, L.: Applications of deep learning to ocean data inference
  and subgrid parameterization, Journal of Advances in Modeling Earth Systems,
  11, 376--399, 2019.

\bibitem[{Brenowitz and Bretherton(2018)}]{brenowitz2018prognostic}
Brenowitz, N.~D. and Bretherton, C.~S.: Prognostic validation of a neural
  network unified physics parameterization, Geophysical Research Letters, 45,
  6289--6298, 2018.

\bibitem[{Brenowitz and Bretherton(2019)}]{brenowitz2019spatially}
Brenowitz, N.~D. and Bretherton, C.~S.: Spatially extended tests of a neural
  network parametrization trained by coarse-graining, Journal of Advances in
  Modeling Earth Systems, 11, 2728--2744, 2019.

\bibitem[{Buizza et~al.(1999)Buizza, Milleer, and
  Palmer}]{buizza1999stochastic}
Buizza, R., Milleer, M., and Palmer, T.~N.: Stochastic representation of model
  uncertainties in the ECMWF ensemble prediction system, Quarterly Journal of
  the Royal Meteorological Society, 125, 2887--2908, 1999.

\bibitem[{Chattopadhyay et~al.(2020{\natexlab{a}})Chattopadhyay, Hassanzadeh,
  and Subramanian}]{chattopadhyay2020data}
Chattopadhyay, A., Hassanzadeh, P., and Subramanian, D.: Data-driven
  predictions of a multiscale Lorenz 96 chaotic system using machine-learning
  methods: reservoir computing, artificial neural network, and long short-term
  memory network, Nonlinear Processes in Geophysics, 27, 373--389,
  2020{\natexlab{a}}.

\bibitem[{Chattopadhyay et~al.(2020{\natexlab{b}})Chattopadhyay, Subel, and
  Hassanzadeh}]{chattopadhyay2020sp}
Chattopadhyay, A., Subel, A., and Hassanzadeh, P.: Data-driven
  super-parameterization using deep learning: Experimentation with multiscale
  Lorenz 96 systems and transfer learning, Journal of Advances in Modeling
  Earth Systems, 12, e2020MS002\,084, 2020{\natexlab{b}}.

\bibitem[{Cho et~al.(2014)Cho, Van~Merri\"enboer, Gulcehre, Bahdanau, Bougares,
  Schwenk, and Bengio}]{gru}
Cho, K., Van~Merri\"enboer, B., Gulcehre, C., Bahdanau, D., Bougares, F.,
  Schwenk, H., and Bengio, Y.: Learning phrase representations using RNN
  encoder-decoder for statistical machine translation, arXiv preprint
  arXiv:1406.1078, 2014.

\bibitem[{Christensen et~al.(2015)Christensen, Moroz, and
  Palmer}]{christensen_regimes}
Christensen, H.~M., Moroz, I.~M., and Palmer, T.~N.: Simulating weather
  regimes: Impact of stochastic and perturbed parameter schemes in a simple
  atmospheric model, Climate Dynamics, 44, 2195--2214, 2015.

\bibitem[{Christensen et~al.(2017)Christensen, Berner, Coleman, and
  Palmer}]{christensen2017stochastic}
Christensen, H.~M., Berner, J., Coleman, D.~R., and Palmer, T.: Stochastic
  parameterization and El Ni{\~n}o--Southern Oscillation, Journal of Climate,
  30, 17--38, 2017.

\bibitem[{Chung et~al.(2016)Chung, Ahn, and Bengio}]{chung2016hierarchical}
Chung, J., Ahn, S., and Bengio, Y.: Hierarchical multiscale recurrent neural
  networks, arXiv preprint arXiv:1609.01704, 2016.

\bibitem[{Crommelin and Vanden-Eijnden(2008)}]{crommelin2008subgrid}
Crommelin, D. and Vanden-Eijnden, E.: Subgrid-scale parameterization with
  conditional Markov chains, Journal of the Atmospheric Sciences, 65,
  2661--2675, 2008.

\bibitem[{Dawson and Palmer(2015)}]{dawson2015simulating}
Dawson, A. and Palmer, T.: Simulating weather regimes: Impact of model
  resolution and stochastic parameterization, Climate Dynamics, 44, 2177--2193,
  2015.

\bibitem[{Eck and Schmidhuber(2002)}]{eck2002first}
Eck, D. and Schmidhuber, J.: A first look at music composition using lstm
  recurrent neural networks, Istituto Dalle Molle Di Studi Sull Intelligenza
  Artificiale, 103, 48, 2002.

\bibitem[{Esteban et~al.(2017)Esteban, Hyland, and
  R{\"a}tsch}]{esteban2017real}
Esteban, C., Hyland, S.~L., and R{\"a}tsch, G.: Real-valued (medical) time
  series generation with recurrent conditional gans, arXiv preprint
  arXiv:1706.02633, 2017.

\bibitem[{Gagne et~al.(2020)Gagne, Christensen, Subramanian, and
  Monahan}]{gan_hannah}
Gagne, D.~J., Christensen, H.~M., Subramanian, A.~C., and Monahan, A.~H.:
  Machine learning for stochastic parameterization: Generative adversarial
  networks in the Lorenz'96 model, Journal of Advances in Modeling Earth
  Systems, 12, e2019MS001\,896, 2020.

\bibitem[{Gentine et~al.(2018)Gentine, Pritchard, Rasp, Reinaudi, and
  Yacalis}]{gentine}
Gentine, P., Pritchard, M., Rasp, S., Reinaudi, G., and Yacalis, G.: Could
  machine learning break the convection parameterization deadlock?, Geophysical
  Research Letters, 45, 5742--5751, 2018.

\bibitem[{Goodfellow et~al.(2014)Goodfellow, Pouget-Abadie, Mirza, Xu,
  Warde-Farley, Ozair, Courville, and Bengio}]{gan_goodfellow}
Goodfellow, I.~J., Pouget-Abadie, J., Mirza, M., Xu, B., Warde-Farley, D.,
  Ozair, S., Courville, A., and Bengio, Y.: Generative adversarial networks,
  arXiv preprint arXiv:1406.2661, 2014.

\bibitem[{Graves(2013)}]{graves2013generating}
Graves, A.: Generating sequences with recurrent neural networks, arXiv preprint
  arXiv:1308.0850, 2013.

\bibitem[{Grover et~al.(2018)Grover, Dhar, and Ermon}]{flow_gan}
Grover, A., Dhar, M., and Ermon, S.: Flow-gan: Combining maximum likelihood and
  adversarial learning in generative models, in: Proceedings of the AAAI
  Conference on Artificial Intelligence, vol.~32, 2018.

\bibitem[{Guillaumin and Zanna(2021)}]{guillaumin2021stochastic}
Guillaumin, A.~P. and Zanna, L.: Stochastic-deep learning parameterization of
  ocean momentum forcing, Journal of Advances in Modeling Earth Systems, 13,
  e2021MS002\,534, 2021.

\bibitem[{Gulrajani et~al.(2017)Gulrajani, Ahmed, Arjovsky, Dumoulin, and
  Courville}]{gulrajani2017improved}
Gulrajani, I., Ahmed, F., Arjovsky, M., Dumoulin, V., and Courville, A.:
  Improved training of wasserstein gans, arXiv preprint arXiv:1704.00028, 2017.

\bibitem[{Hasselmann(1976)}]{hasselmann1976stochastic}
Hasselmann, K.: Stochastic climate models part I. Theory, tellus, 28, 473--485,
  1976.

\bibitem[{Hochreiter and Schmidhuber(1997)}]{lstm}
Hochreiter, S. and Schmidhuber, J.: Long short-term memory, Neural computation,
  9, 1735--1780, 1997.

\bibitem[{Johnson et~al.(2019)Johnson, Stockdale, Ferranti, Balmaseda, Molteni,
  Magnusson, Tietsche, Decremer, Weisheimer, Balsamo et~al.}]{johnson2019seas5}
Johnson, S.~J., Stockdale, T.~N., Ferranti, L., Balmaseda, M.~A., Molteni, F.,
  Magnusson, L., Tietsche, S., Decremer, D., Weisheimer, A., Balsamo, G.,
  et~al.: SEAS5: the new ECMWF seasonal forecast system, Geoscientific Model
  Development, 12, 1087--1117, 2019.

\bibitem[{Kingma and Ba(2014)}]{kingma2014adam}
Kingma, D.~P. and Ba, J.: Adam: A method for stochastic optimization, arXiv
  preprint arXiv:1412.6980, 2014.

\bibitem[{Kingma and Welling(2013)}]{vae}
Kingma, D.~P. and Welling, M.: Auto-encoding variational bayes, arXiv preprint
  arXiv:1312.6114, 2013.

\bibitem[{Krasnopolsky et~al.(2013)Krasnopolsky, Fox-Rabinovitz, and
  Belochitski}]{krasnopolsky}
Krasnopolsky, V.~M., Fox-Rabinovitz, M.~S., and Belochitski, A.~A.: Using
  ensemble of neural networks to learn stochastic convection parameterizations
  for climate and numerical weather prediction models from data simulated by a
  cloud resolving model, Advances in Artificial Neural Systems, 2013, 2013.

\bibitem[{Kwasniok(2012)}]{kwasniok2012data}
Kwasniok, F.: Data-based stochastic subgrid-scale parametrization: an approach
  using cluster-weighted modelling, Philosophical Transactions of the Royal
  Society A: Mathematical, Physical and Engineering Sciences, 370, 1061--1086,
  2012.

\bibitem[{Leutbecher and Palmer(2008)}]{leutbecher2008ensemble}
Leutbecher, M. and Palmer, T.~N.: Ensemble forecasting, Journal of
  computational physics, 227, 3515--3539, 2008.

\bibitem[{Leutbecher et~al.(2017)Leutbecher, Lock, Ollinaho, Lang, Balsamo,
  Bechtold, Bonavita, Christensen, Diamantakis, Dutra
  et~al.}]{leutbecher2017stochastic}
Leutbecher, M., Lock, S.-J., Ollinaho, P., Lang, S.~T., Balsamo, G., Bechtold,
  P., Bonavita, M., Christensen, H.~M., Diamantakis, M., Dutra, E., et~al.:
  Stochastic representations of model uncertainties at ECMWF: State of the art
  and future vision, Quarterly Journal of the Royal Meteorological Society,
  143, 2315--2339, 2017.

\bibitem[{Liu et~al.(2020)Liu, Kutz, and Brunton}]{liu2020hierarchical}
Liu, Y., Kutz, J.~N., and Brunton, S.~L.: Hierarchical deep learning of
  multiscale differential equation time-steppers, arXiv preprint
  arXiv:2008.09768, 2020.

\bibitem[{Lorenz(1996)}]{lorenz1996predictability}
Lorenz, E.~N.: Predictability: A problem partly solved, in: Proc. Seminar on
  predictability, vol.~1, 1996.

\bibitem[{Lorenz(2006)}]{lorenz2006regimes}
Lorenz, E.~N.: Regimes in simple systems, Journal of the atmospheric sciences,
  63, 2056--2073, 2006.

\bibitem[{Mogren(2016)}]{mogren2016c}
Mogren, O.: C-RNN-GAN: Continuous recurrent neural networks with adversarial
  training, arXiv preprint arXiv:1611.09904, 2016.

\bibitem[{Molteni et~al.(2011)Molteni, Stockdale, Balmaseda, Balsamo, Buizza,
  Ferranti, Magnusson, Mogensen, Palmer, and Vitart}]{molteni2011new}
Molteni, F., Stockdale, T., Balmaseda, M., Balsamo, G., Buizza, R., Ferranti,
  L., Magnusson, L., Mogensen, K., Palmer, T., and Vitart, F.: The new ECMWF
  seasonal forecast system (System 4), vol.~49, European Centre for
  medium-range weather forecasts Reading, 2011.

\bibitem[{O'Gorman and Dwyer(2018)}]{ogorman_2018}
O'Gorman, P.~A. and Dwyer, J.~G.: Using machine learning to parameterize moist
  convection: Potential for modeling of climate, climate change, and extreme
  events, Journal of Advances in Modeling Earth Systems, 10, 2548--2563, 2018.

\bibitem[{Palmer(2012)}]{palmer2012stochastic}
Palmer, T.~N.: Towards the probabilistic Earth-system simulator: A vision for
  the future of climate and weather prediction, Quarterly Journal of the Royal
  Meteorological Society, 138, 841--861, 2012.

\bibitem[{Palmer et~al.(2009)Palmer, Buizza, Doblas-Reyes, Jung, Leutbecher,
  Shutts, Steinheimer, and Weisheimer}]{palmer2009stochastic}
Palmer, T.~N., Buizza, R., Doblas-Reyes, F., Jung, T., Leutbecher, M., Shutts,
  G.~J., Steinheimer, M., and Weisheimer, A.: Stochastic parametrization and
  model uncertainty, 2009.

\bibitem[{Rasp(2020)}]{rasp2020}
Rasp, S.: Coupled online learning as a way to tackle instabilities and biases
  in neural network parameterizations: general algorithms and Lorenz 96 case
  study (v1. 0), Geoscientific Model Development, 13, 2185--2196, 2020.

\bibitem[{Rasp et~al.(2018)Rasp, Pritchard, and Gentine}]{rasp2018deep}
Rasp, S., Pritchard, M.~S., and Gentine, P.: Deep learning to represent subgrid
  processes in climate models, Proceedings of the National Academy of Sciences,
  115, 9684--9689, 2018.

\bibitem[{Sanchez et~al.(2016)Sanchez, Williams, and
  Collins}]{sanchez2016improved}
Sanchez, C., Williams, K.~D., and Collins, M.: Improved stochastic physics
  schemes for global weather and climate models, Quarterly Journal of the Royal
  Meteorological Society, 142, 147--159, 2016.

\bibitem[{Schneider et~al.(2017)Schneider, Teixeira, Bretherton, Brient,
  Pressel, Sch{\"a}r, and Siebesma}]{schneider_clouds}
Schneider, T., Teixeira, J., Bretherton, C.~S., Brient, F., Pressel, K.~G.,
  Sch{\"a}r, C., and Siebesma, A.~P.: Climate goals and computing the future of
  clouds, Nature Climate Change, 7, 3--5, 2017.

\bibitem[{Selten and Branstator(2004)}]{selten2004preferred}
Selten, F.~M. and Branstator, G.: Preferred regime transition routes and
  evidence for an unstable periodic orbit in a baroclinic model, Journal of the
  atmospheric sciences, 61, 2267--2282, 2004.

\bibitem[{Skamarock et~al.(2019)Skamarock, Klemp, Dudhia, Gill, Liu, Berner,
  Wang, Powers, Duda, Barker et~al.}]{skamarock2019description}
Skamarock, W.~C., Klemp, J.~B., Dudhia, J., Gill, D.~O., Liu, Z., Berner, J.,
  Wang, W., Powers, J.~G., Duda, M.~G., Barker, D.~M., et~al.: A description of
  the advanced research WRF model version 4, National Center for Atmospheric
  Research: Boulder, CO, USA, 145, 145, 2019.

\bibitem[{Stephenson et~al.(2004)Stephenson, Hannachi, and
  O'Neill}]{stephenson2004existence}
Stephenson, D., Hannachi, A., and O'Neill, A.: On the existence of multiple
  climate regimes, Quarterly Journal of the Royal Meteorological Society: A
  journal of the atmospheric sciences, applied meteorology and physical
  oceanography, 130, 583--605, 2004.

\bibitem[{Stockdale et~al.(2011)Stockdale, Anderson, Balmaseda, Doblas-Reyes,
  Ferranti, Mogensen, Palmer, Molteni, and Vitart}]{stockdale2011ecmwf}
Stockdale, T.~N., Anderson, D.~L., Balmaseda, M.~A., Doblas-Reyes, F.,
  Ferranti, L., Mogensen, K., Palmer, T.~N., Molteni, F., and Vitart, F.: ECMWF
  seasonal forecast system 3 and its prediction of sea surface temperature,
  Climate dynamics, 37, 455--471, 2011.

\bibitem[{Straus et~al.(2007)Straus, Corti, and
  Molteni}]{straus2007circulation}
Straus, D.~M., Corti, S., and Molteni, F.: Circulation regimes: Chaotic
  variability versus SST-forced predictability, Journal of climate, 20,
  2251--2272, 2007.

\bibitem[{Sutskever et~al.(2011)Sutskever, Martens, and
  Hinton}]{sutskever2011generating}
Sutskever, I., Martens, J., and Hinton, G.~E.: Generating text with recurrent
  neural networks, in: ICML, 2011.

\bibitem[{Sutskever et~al.(2014)Sutskever, Vinyals, and
  Le}]{sutskever2014sequence}
Sutskever, I., Vinyals, O., and Le, Q.~V.: Sequence to sequence learning with
  neural networks, in: Advances in neural information processing systems, pp.
  3104--3112, 2014.

\bibitem[{Theis et~al.(2015)Theis, Oord, and Bethge}]{eval_generative_models}
Theis, L., Oord, A. v.~d., and Bethge, M.: A note on the evaluation of
  generative models, arXiv preprint arXiv:1511.01844, 2015.

\bibitem[{Vaswani et~al.(2017)Vaswani, Shazeer, Parmar, Uszkoreit, Jones,
  Gomez, Kaiser, and Polosukhin}]{vaswani2017attention}
Vaswani, A., Shazeer, N., Parmar, N., Uszkoreit, J., Jones, L., Gomez, A.~N.,
  Kaiser, {\L}., and Polosukhin, I.: Attention is all you need, in: Advances in
  neural information processing systems, pp. 5998--6008, 2017.

\bibitem[{Vlachas et~al.(2018)Vlachas, Byeon, Wan, Sapsis, and
  Koumoutsakos}]{vlachas2018data}
Vlachas, P.~R., Byeon, W., Wan, Z.~Y., Sapsis, T.~P., and Koumoutsakos, P.:
  Data-driven forecasting of high-dimensional chaotic systems with long
  short-term memory networks, Proceedings of the Royal Society A: Mathematical,
  Physical and Engineering Sciences, 474, 20170\,844, 2018.

\bibitem[{Vlachas et~al.(2020)Vlachas, Pathak, Hunt, Sapsis, Girvan, Ott, and
  Koumoutsakos}]{vlachas2020backpropagation}
Vlachas, P.~R., Pathak, J., Hunt, B.~R., Sapsis, T.~P., Girvan, M., Ott, E.,
  and Koumoutsakos, P.: Backpropagation algorithms and reservoir computing in
  recurrent neural networks for the forecasting of complex spatiotemporal
  dynamics, Neural Networks, 126, 191--217, 2020.

\bibitem[{Vlachas et~al.(2022)Vlachas, Arampatzis, Uhler, and
  Koumoutsakos}]{vlachas2022multiscale}
Vlachas, P.~R., Arampatzis, G., Uhler, C., and Koumoutsakos, P.: Multiscale
  simulations of complex systems by learning their effective dynamics, Nature
  Machine Intelligence, 4, 359--366, 2022.

\bibitem[{Walters et~al.(2019)Walters, Baran, Boutle, Brooks, Earnshaw,
  Edwards, Furtado, Hill, Lock, Manners et~al.}]{walters2019met}
Walters, D., Baran, A.~J., Boutle, I., Brooks, M., Earnshaw, P., Edwards, J.,
  Furtado, K., Hill, P., Lock, A., Manners, J., et~al.: The Met Office Unified
  Model global atmosphere 7.0/7.1 and JULES global land 7.0 configurations,
  Geoscientific Model Development, 12, 1909--1963, 2019.

\bibitem[{Yuval and O’Gorman(2020)}]{yuval_rf}
Yuval, J. and O’Gorman, P.~A.: Stable machine-learning parameterization of
  subgrid processes for climate modeling at a range of resolutions, Nature
  communications, 11, 1--10, 2020.

\bibitem[{Yuval et~al.(2021)Yuval, O'Gorman, and Hill}]{yuval2021}
Yuval, J., O'Gorman, P.~A., and Hill, C.~N.: Use of neural networks for stable,
  accurate and physically consistent parameterization of subgrid atmospheric
  processes with good performance at reduced precision, Geophysical Research
  Letters, 48, e2020GL091\,363, 2021.

\bibitem[{Zelinka et~al.(2020)Zelinka, Myers, McCoy, Po-Chedley, Caldwell,
  Ceppi, Klein, and Taylor}]{zelinka2020causes}
Zelinka, M.~D., Myers, T.~A., McCoy, D.~T., Po-Chedley, S., Caldwell, P.~M.,
  Ceppi, P., Klein, S.~A., and Taylor, K.~E.: Causes of higher climate
  sensitivity in CMIP6 models, Geophysical Research Letters, 47,
  e2019GL085\,782, 2020.

\bibitem[{Zhao et~al.(2020)Zhao, Cong, Dai, and Carin}]{zhao2020bridging}
Zhao, M., Cong, Y., Dai, S., and Carin, L.: Bridging Maximum Likelihood and
  Adversarial Learning via $\alpha$-Divergence, in: Proceedings of the AAAI
  Conference on Artificial Intelligence, vol.~34, pp. 6901--6908, 2020.

\end{thebibliography}

\end{document}